%% file: main.tex
\documentclass[lettersize,journal]{IEEEtran}
\usepackage{amsmath,amsfonts}
\usepackage{algorithmic}
\usepackage{algorithm}
\usepackage{booktabs}
\usepackage{multirow}
\usepackage{array}
\usepackage[caption=false,font=normalsize,labelfont=sf,textfont=sf]{subfig}
\usepackage{textcomp}
\usepackage{stfloats}
\usepackage{url}
\usepackage{verbatim}
\usepackage{graphicx}
\usepackage{enumitem}
\usepackage{cite}
\usepackage[bookmarks=true]{hyperref}
\usepackage{cleveref}
\usepackage{orcidlink}
\usepackage{xcolor}
\usepackage{xspace}
\usepackage{pifont}  
\input{format/macros}

\usepackage{threeparttable} 
\usepackage[table]{xcolor} 

\begin{document}

\title{\LARGE ObjSplat: Geometry-Aware Gaussian Surfels for\\ Active Object Reconstruction}

\markboth{IEEE Transactions on Automation Science and Engineering. Preprint Version. Accepted May, 2026}
{Li \MakeLowercase{\textit{et al.}}: ActiveSplat} 

\author{Yuetao Li, Zhizhou Jia, Yu Zhang, Qun Hao, and Shaohui Zhang
\thanks{This work was supported in part by the National Key Research and Development Program of China under Grant 2021YFC2202404 and in part by the National Natural Science Foundation of China under Grant 62275020. \textit{(Corresponding author: Shaohui Zhang.)}}
\thanks{Yuetao Li, Zhizhou Jia, Yu Zhang, and Shaohui Zhang are with the School of Optics and Photonics, Beijing Institute of Technology, Beijing 100081, China. (e-mail: liyuetaochn@bit.edu.cn; jiazhizhou@bit.edu.cn; zhangyu\_bit@bit.edu.cn; zhangshaohui@bit.edu.cn).}
\thanks{Qun Hao is with the School of Optics and Photonics, Beijing Institute of Technology, Beijing 100081, China, and also with the School of Optoelectronic Engineering, Changchun University of Science and Technology, Changchun 130022, China (e-mail: qhao@bit.edu.cn).}
\thanks{Digital Object Identifier (DOI): 10.1109/TASE.2026.3700105.}
\thanks{Project page: \url{https://li-yuetao.github.io/ObjSplat-page/}.}
}


\maketitle

\begin{abstract}
Autonomous high-fidelity object reconstruction is fundamental for creating digital assets and bridging the simulation-to-reality gap in robotics. We present ObjSplat, an active reconstruction framework that leverages Gaussian surfels as a unified representation to progressively reconstruct unknown objects with both photorealistic appearance and accurate geometry. Addressing the limitations of conventional opacity or depth-based cues, we introduce a geometry-aware viewpoint evaluation pipeline that explicitly models back-face visibility and occlusion-aware multi-view covisibility, reliably identifying under-reconstructed regions even on geometrically complex objects. Furthermore, to overcome the limitations of greedy planning strategies, ObjSplat employs a next-best-path (NBP) planner that performs multi-step lookahead on a dynamically constructed spatial graph. By jointly optimizing information gain and movement cost, this planner generates globally efficient trajectories. Extensive experiments in simulation and on real-world cultural artifacts demonstrate that ObjSplat produces physically consistent models within minutes, achieving superior reconstruction fidelity and surface completeness while significantly reducing scan time and path length compared to state-of-the-art approaches.
\end{abstract}

\def\abstractname{Note to Practitioners}
\begin{abstract}
This paper addresses the challenge of autonomous, high-fidelity digitization of physical objects, a capability essential for digital cultural heritage preservation and XR asset creation. Currently, existing automated systems often rely on pre-programmed trajectories that cannot adapt to unknown shapes or utilize local planning strategies that result in inefficient, redundant movements. We present ObjSplat, a unified active reconstruction system that overcomes the inefficiencies of manual scanning and the limitations of existing predefined trajectories or greedy automation methods. By leveraging Gaussian surfels and a geometry-aware evaluation pipeline, the system reliably identifies under-reconstructed regions on complex objects (e.g., hollow or thin structures). Unlike traditional view-by-view strategies, our multi-step Next-Best-Path (NBP) planner optimizes global movement, significantly reducing operation time and redundant motion. The framework is ready for deployment on a robotic arm equipped with RGB-D sensors to produce physically consistent, watertight models within minutes, with future potential to address complex optical properties and multi-robot collaboration.
\end{abstract}


\input{src/1_introduction}
\input{src/2_related_works}
\input{src/3_method}
\input{src/4_experiments}
\input{src/5_conclusion}

\bibliographystyle{IEEEtran}
\bibliography{citations}

\vfill

\end{document}

%% file: format/macros.tex
\definecolor{MyDarkBlue}{rgb}{0,0.08,1}
\definecolor{airforceblue}{rgb}{0.36, 0.54, 0.66}
\definecolor{MyDarkGreen}{rgb}{0.02,0.6,0.02}
\definecolor{MyDarkRed}{rgb}{0.8,0.02,0.02}
\definecolor{MyDarkOrange}{rgb}{0.40,0.2,0.02}
\definecolor{MyPurple}{RGB}{111,0,255}
\definecolor{MyRed}{rgb}{1.0,0.0,0.0}
\definecolor{MyGold}{rgb}{0.75,0.6,0.12}
\definecolor{MyDarkgray}{rgb}{0.66, 0.66, 0.66}
\definecolor{MyPink}{rgb}{0.9, 0.33, 0.5}
\definecolor{MyCyan}{rgb}{0., 0.4, 0.4}
\definecolor{MyBlue}{rgb}{0.5, 0.8, 0.9}
\definecolor{AbsoluteColor}{rgb}{0.76, 0.2, 0.2}
\definecolor{DeltaColor}{rgb}{0.87, 0.72, 0.3}
\definecolor{RelativeColor}{rgb}{0.04, 0.33, 0.58}
\definecolor{StanfordRed}{rgb}{0.549, 0.082, 0.082}
\definecolor{BITBrown}{HTML}{A13E0B}
\definecolor{BITGreen}{HTML}{006C39}
\definecolor{deepred2}{RGB}{196, 49, 25}

\crefname{section}{Sec.}{Secs.}
\crefname{figure}{Fig.}{Figs.}
\crefname{table}{Tab.}{Tabs.}
\crefname{equation}{Eq.}{Eqs.}
\Crefname{section}{Section}{Sections}
\Crefname{table}{Table}{Tables}

\newcommand{\ours}[0]{ObjSplat\xspace}

\definecolor{deepgreen}{RGB}{63, 126, 49}
\definecolor{deepred2}{RGB}{196, 49, 25}


%% file: src/1_introduction.tex
\section{Introduction}
\label{sec:introduction}
\IEEEPARstart{H}{igh-quality} digitization of physical objects, characterized by both accurate geometry and photorealistic appearance, has been a long-standing pursuit in robotics and computer vision\cite{yu2025metascenes}. Such fine-grained digital models are fundamental to creating immersive extended reality (XR) experiences and ensuring the long-term preservation of digital cultural heritage\cite{dong2025digital}. Notably, in the robotics domain, these models hold the potential to bridge the simulation-to-reality (sim-to-real) gap, empowering the robust development and validation of perception, planning, and control algorithms within physics-based simulated environments~\cite{deng2025best, torne2024reconciling, pfaff2025scalable, yu2025real2render2real, escontrela2025gaussgym}.

\begin{figure}[t]
    \centering
    \includegraphics[width=1.0\linewidth]{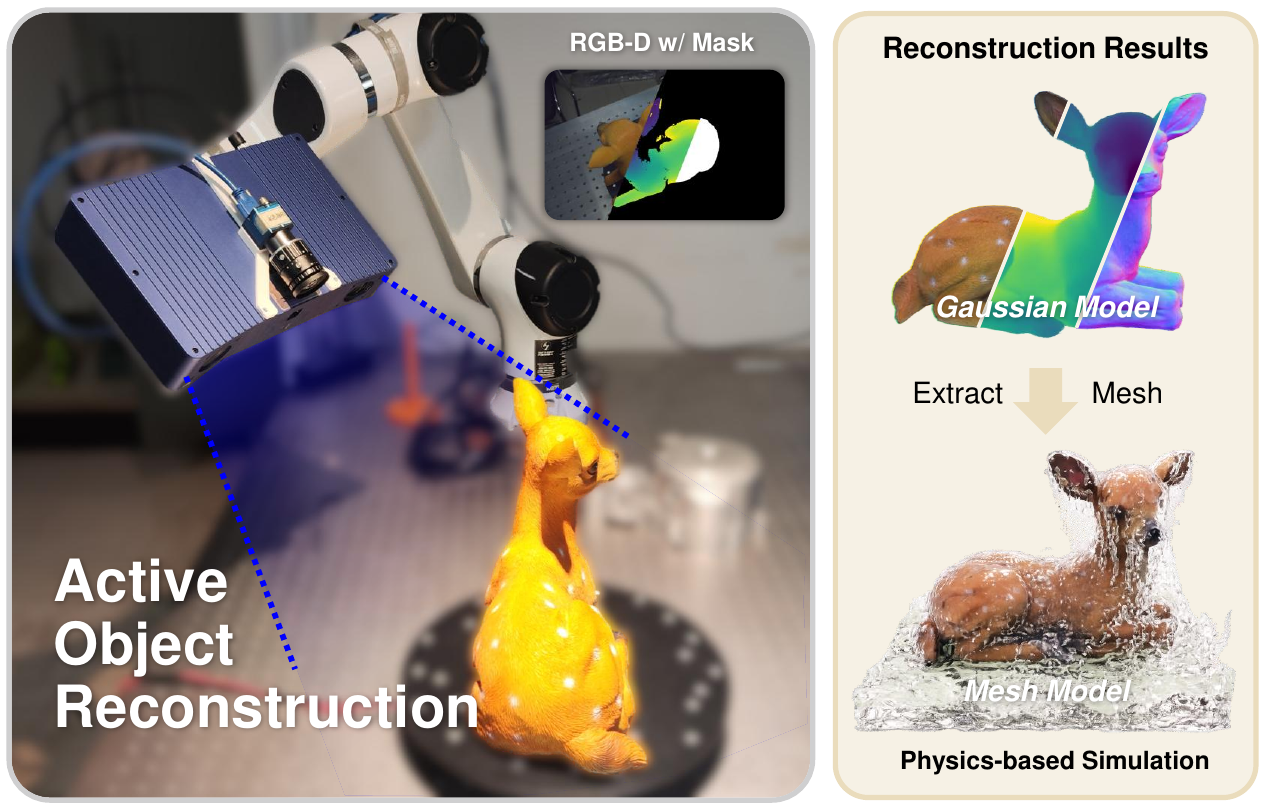}
    \caption{ObjSplat autonomously plans viewpoints and progressively reconstructs an unknown object into a high-fidelity Gaussian model and water-tight mesh, enabling direct use in physics simulations.}
    \label{fig: teaser}
    \vspace{-6mm}
\end{figure}

Recent advances in radiance fields, represented by neural radiance fields (NeRF)~\cite{mildenhall2021nerf} and 3D Gaussian splatting (3DGS)~\cite{kerbl20233d}, have demonstrated remarkable promise in high-quality reconstruction and novel view synthesis. However, achieving these results typically relies on dense, passive data acquisition and complex post-processing pipelines involving structure-from-motion (SfM)~\cite{schoenberger2016sfm} initialization and computationally intensive offline optimization. This often requires substantial manual intervention, fundamentally limiting their scalable application in object-level digital asset creation.
To mitigate this issue, active object reconstruction (AOR) has emerged as a promising solution, aiming to automate the reconstruction process with robotic systems. The core of this task lies in accurately identifying under-reconstructed regions (e.g., incomplete or poor-quality areas) and planning a sequence of viewpoints for objects with diverse geometries and appearances, aiming to achieve complete reconstruction efficiently. Many well-established approaches formulate this task as a sequential next-best-view (NBV) planning problem~\cite{lee2022uncertainty, pan2022activenerf, yan2023active, jin2023neu, border2024surface, ye2024pvp, jiang2024fisherrf, xue2024neural, chen2024gennbv, xie2025gauss, jia2025pb, lee2025bayesian}, where methods leverage cues such as frontier information~\cite{border2024surface, jia2025pb}, color and density distributions~\cite{pan2022activenerf}, or visibility fields~\cite{xue2024neural} to quantify the uncertainty or information gain of candidate views.  While effective in identifying informative views, traditional iterative NBV strategies are often limited to single-step optimization, ignoring the robot's kinematic constraints and movement costs, which often results in lengthy and suboptimal scanning trajectories. In parallel, learning-based approaches such as reinforcement learning (RL) policies~\cite{peralta2020next, chen2025gleam, ye2026efficient} can implicitly capture long-horizon objectives and optimize movement costs~\cite{li2025nextbestpath, chen2025gleam, dobivs2025next}. While some works further mitigate this issue by introducing one-shot view planners~\cite{pan2022scvp, pan2024many, pan2024integrating, pan2024exploiting, pan2025dm}, a fundamental challenge persists: \emph{how to reliably assess reconstruction completeness and quality, and critically, how to strike an effective balance between efficient exploration and high-quality reconstruction}.

In this work, we introduce \textbf{ObjSplat}, an active object reconstruction system leveraging 2D Gaussian surfels as a unified representation for both model optimization and exploration guidance. Our system is designed to autonomously and progressively achieve high-fidelity geometric and textural reconstruction from RGB-D data, producing physically consistent assets ready for downstream tasks (see Fig.~\ref{fig: teaser}). The core of our approach lies in tightly coupling high-quality reconstruction with efficient exploration within a unified framework, enabling both accuracy and efficiency in object-level reconstruction. Specifically, we leverage surface normals to guide the incremental update of the Gaussian model and facilitate joint geometry–texture optimization. To guide the active reconstruction of unknown objects, we design a geometry-aware viewpoint evaluation strategy. Unlike existing methods that primarily rely on opacity or depth residuals to detect under-reconstructed regions \cite{matsuki2024gaussian, li2025activesplat}, our strategy specifically addresses object-centric viewpoints and non-closed surfaces, enabling more accurate detection of incomplete and unobserved back-face regions as well as robust quantification of true multi-view covisibility, which is essential for handling challenging hollow, self-occluded, and thin structures. On the exploration side, we move beyond the traditional greedy NBV paradigm. Instead, we propose a next-best-path (NBP) planning strategy, which performs multi-step lookahead planning on a dynamically constructed spatial topology, jointly accounting for information gain and movement cost to generate globally superior scanning trajectories while significantly reducing redundant movements and online scan time. Both simulation and real-world experiments demonstrate that our system adaptively achieves complete, high-fidelity reconstruction of unknown objects (with varying geometric and textural complexities) with fewer movements, consistently outperforming state-of-the-art baselines in both efficiency and reconstruction quality.

In summary, our work advances the state of the art in active object reconstruction through the following contributions:

\begin{itemize}
    \item A geometry-aware view evaluation pipeline that accurately quantifies reconstruction quality and completeness for non-closed surfaces, by explicitly modeling back-face visibility and true multi-view covisibility, providing reliable guidance for both exploration and refinement.
    \item A next-best-path (NBP) planning strategy that performs multi-step lookahead on a dynamically constructed spatial graph, jointly optimizing information gain and movement cost to generate efficient scanning trajectories, significantly reducing total path length and scan time compared to conventional greedy NBV approaches.
    \item A unified active object reconstruction framework, ObjSplat, which leverages surface normals for the incremental update of Gaussian surfels and joint geometry–texture optimization. This tight coupling enables the efficient, autonomous production of physically consistent, high-fidelity digital assets ready for downstream tasks.
    \item Extensive experiments on objects with diverse geometric and textural complexities demonstrate that ObjSplat consistently outperforms existing methods in both reconstruction quality and exploration efficiency, while real-world experiments on four cultural heritage artifacts further validate its robustness and practical applicability.
\end{itemize}

%% file: src/2_related_works.tex
\section{Related Work}
\label{sec:related_work}
Our work builds upon two primary research areas: high-quality reconstruction and active view planning for robotics. In this section, we review the key advancements in both fields.

\subsection{High-Quality Reconstruction}
\label{subsec:high_quality_recon}
High-quality reconstruction aims to achieve both accurate geometry and photorealistic appearance. Classical pipelines primarily rely on SfM/MVS~\cite{schoenberger2016mvs}, or the fusion of high-precision depth measurements (e.g., structured light) to estimate dense point clouds. These raw representations are subsequently converted into explicit surface meshes via TSDF fusion or Poisson reconstruction~\cite{kazhdan2013screened}, followed by texture mapping~\cite{fu2021seamless, zhang2022adaptive}. While these explicit methods offer metrically consistent surfaces and benefit from mature engineering ecosystems, they often require dense, well-conditioned observations and tend to degrade on texture-poor or severely occluded objects, with limited capability in reproducing high-fidelity view-dependent appearance. Subsequently, NeRF~\cite{mildenhall2021nerf, muller2022instant} and 3DGS~\cite{kerbl20233d} have demonstrated the powerful capability of radiance field representation in synthesizing high-fidelity novel views. To further improve geometric quality, subsequent methods have integrated signed distance functions (SDFs)~\cite{wang2021neus} and multi-resolution 3D hash grids~\cite{li2023neuralangelo}, and other techniques~\cite{zhang2025nerfprior, han2025sparserecon} to generate more accurate surface reconstructions. However, NeRF-based methods often suffer from high computational costs and rendering inefficiencies due to their implicit nature. In contrast, recent GS-based follow-up works enhance geometric fidelity while reducing computational overhead by incorporating explicit geometry representations and normal priors~\cite{huang20242d, chen2024pgsr, dai2024high, zhang2025quadratic}, as well as applying depth regularization~\cite{liu2025gs, hong2025gs}, making it more efficient for real-time applications. More recently, mesh-based Gaussian splatting methods~\cite{guedon2025milo, Held2025Triangle, Held2025MeshSplatting} have emerged, which directly bind Gaussians to mesh surfaces or utilize restricted Delaunay triangulation to enforce surface consistency, significantly bridging the gap between volumetric rendering and explicit surface reconstruction. In addition, to address the challenges of data scarcity in sparse-view and object-centric scenarios, recent works~\cite{yang2024gaussianobject, fan2024instantsplat, bao2025free360} leverage strong external priors—ranging from diffusion-based generative models~\cite{yang2024gaussianobject, bao2025free360} to geometric foundation models~\cite{fan2024instantsplat}—to infer missing geometry and resolve spatial ambiguities. While these approaches achieve impressive visual fidelity by hallucinating plausible details from minimal inputs, they remain fundamentally passive and may struggle with hallucination artifacts or over-smoothed details on unseen complex objects.

To address this, our approach resolves the information deficiency through active observation, building on a Gaussian-surfel representation, automating the data collection process via a robotic arm and turntable setup. Notably, unlike traditional disjointed multi-view reconstruction pipelines, our system seamlessly integrates online scanning and offline refinement within a unified optimization framework, ensuring consistent model evolution. By introducing a joint geometry–texture optimization strategy, we effectively enhance both global consistency and geometric precision, enabling the autonomous creation of high-fidelity digital twins.

\subsection{Active Object Reconstruction}
\label{subsec:active_object_recon}

Active object reconstruction has emerged as a task due to the limited field of view (FoV) of vision sensors and the inability of predefined scanning trajectories to meet the needs of a wide range of object reconstruction tasks, including those with varying degrees of complex textures (e.g., repetitive or sparse patterns) and complex geometry (e.g., self-occlusion, non-convexity). Active object reconstruction, as the object-level reconstruction task within active vision tasks, aims to autonomously plan a series of viewpoints to construct a complete model of an unknown object efficiently. Existing strategies can be broadly categorized into dominant paradigm progressive iterative planning~\cite{lee2022uncertainty, pan2022activenerf, yan2023active, jin2023neu, border2024surface, ye2024pvp, jiang2024fisherrf, xue2024neural, chen2024gennbv, xie2025gauss, jia2025pb, lee2025bayesian} and global one-shot planning~\cite{pan2022scvp, pan2024many, pan2024integrating, pan2024exploiting, pan2025dm}.

The iterative paradigm sequentially selects a single next viewpoint to maximize an information gain metric. Early approaches focused on geometric completeness, using frontier-based methods~\cite{border2024surface, jia2025pb} to gradually expand the boundaries toward a complete reconstruction. PB-NBV~\cite{jia2025pb} leverages GMMs to fit voxels of varying visibility and employs a projection-based approach to accelerate viewpoint evaluation. Sample-based methods, on the other hand, often require elaborate heuristic sampling strategies (uniform~\cite{xu2025hgs}, random~\cite{jin2025activegs}, or sampling at topology nodes~\cite{li2025activesplat}), and inherent candidate viewpoints are not adaptive for unknown objects of varying geometrical morphology. PVP-Recon~\cite{ye2024pvp} employs a progressive approach, where view selection is guided by warping consistency under sparse view surface reconstruction. With the emergence of radiance fields, recent methods have shifted towards quantifying visual or model uncertainty. For example, ActiveNeRF~\cite{pan2022activenerf} models color distributions, while information-theoretic methods such as FisherRF~\cite{jiang2024fisherrf} and GauSS-MI~\cite{xie2025gauss} leverage Fisher information and Shannon mutual information, respectively, to guide view selection towards regions of high model uncertainty. While these methods are effective in identifying informative views, they often struggle with complex geometries like self-occlusions due to a lack of explicit geometric awareness. Moreover, traditional greedy, single-step optimization ignores movement costs, typically resulting in inefficient scanning trajectories. To address this, recent research has shifted towards non-greedy policies, including learned utility functions~\cite{hepp2018learn} and RL-based strategies~\cite{peralta2020next, chen2025gleam, ye2026efficient} that capture long-horizon objectives. While these approaches improve planning foresight, they often require extensive offline training and may struggle with out-of-distribution (OOD) scenarios or across different hardware setups. Furthermore, some recent approaches balance efficiency and cost~\cite{xu2025hgs, li2025activesplat, li2025nextbestpath}, they are designed for scene-level tasks and fail to address the specific geometric challenges of object reconstruction.

To mitigate the inefficiencies of iterative planning, one-shot methods~\cite{pan2022scvp, pan2024integrating} frame one-shot viewpoint planning as a set covering optimization problem, aiming to obtain the minimum set of viewpoints for dense coverage and solving the traveling salesman problem (TSP) to generate a complete scan trajectory in a single step. With the introduction of diffusion models, powerful geometric priors have been leveraged to enable RGB-based active object reconstruction. For example, DM-OSVP~\cite{pan2024exploiting} and its improved version DM-OSVP++~\cite{pan2025dm} use 3D diffusion priors to generate geometry from RGB data, enabling the selection of a compact, information-rich set of views. DM-OSVP++ further introduces geometric/texture complexity conditioning, which is beneficial for RGB-only planning. However, these methods struggle with unknown scales and severe occlusions due to their reliance on coarse, early-stage priors. Furthermore, the substantial inference latency of diffusion models hinders real-time performance, and their "one-shot" nature lacks the adaptability required to progressively refine reconstruction for complex, unseen objects.

Our work, ObjSplat, effectively bridges the gap between short-sighted NBV planning and one-shot methods. We propose an NBP strategy that performs multi-step lookahead on a spatial topology, optimizing the trade-off between information gain and movement cost. This approach combines the adaptability of progressive reconstruction with the foresight of global planning. Consequently, ObjSplat achieves a superior trade-off, balancing reconstruction quality and exploration efficiency in a unified, progressive planning framework.

%% file: src/3_method.tex
\begin{figure*}[t]
  \centering
  \includegraphics[width=0.95\linewidth]{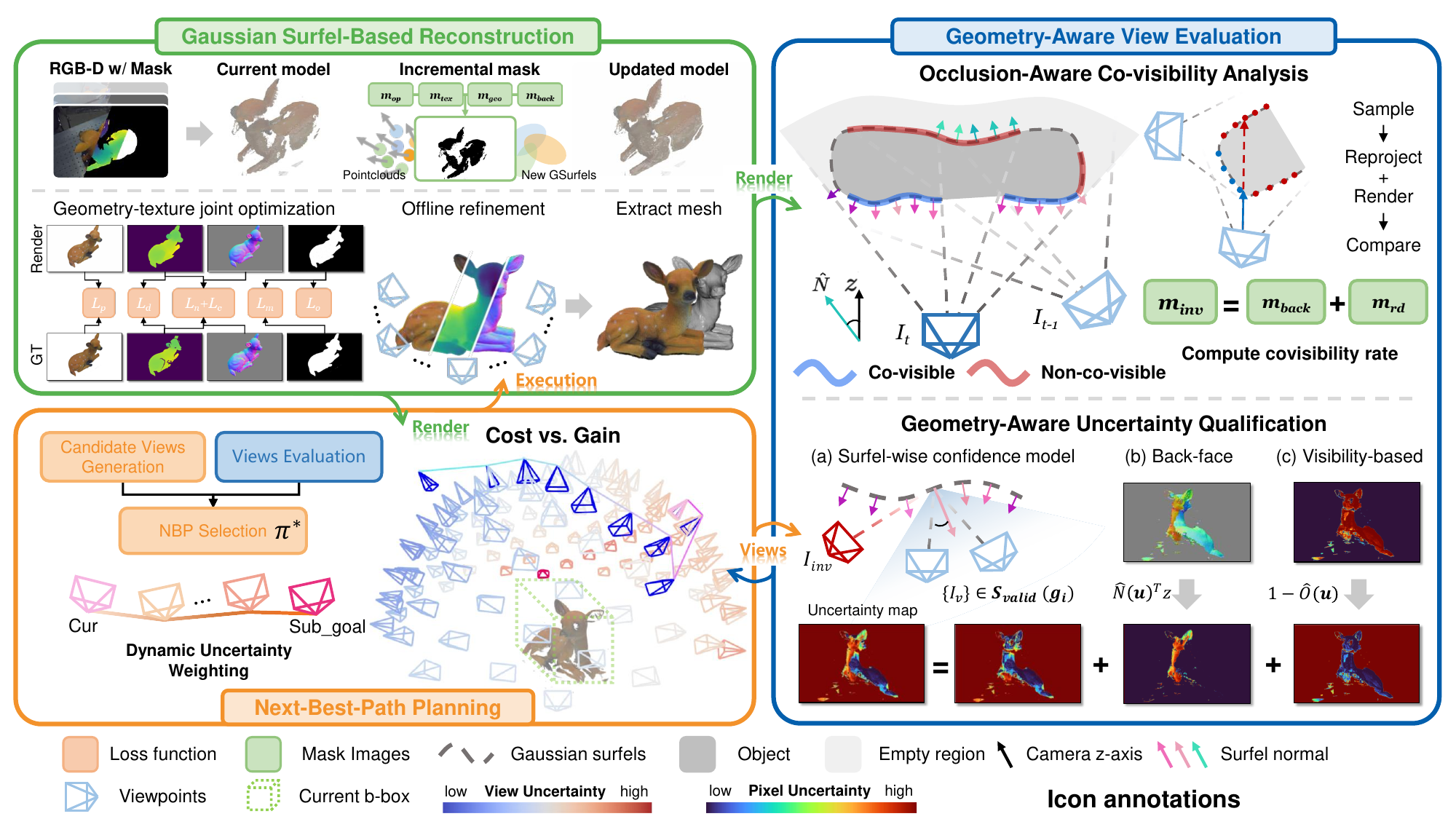}
  \vspace{-2mm}
  \caption{\textbf{System Overview.} \ours progressively reconstructs unknown objects from RGB-D frames using Gaussian surfels as a unified representation. (\textbf{Top left}) Incoming frames are fused into the global model, where geometry–texture joint optimization, enforcing both photometric and geometric consistency. (\textbf{Right}) A geometry-aware view evaluation pipeline renders an uncertainty map by integrating occlusion-aware covisibility, surfel-wise confidence, and back-face detection to quantify surface quality and completeness. (\textbf{Bottom left}) Guided by this dense uncertainty, the next-best-path planner performs multi-step lookahead on a spatial topology, generating globally efficient trajectories that balance information gain and movement cost for active reconstruction.}
  \label{fig:pipeline}
  \vspace{-2mm}
\end{figure*}

\section{Methodology}
\label{sec:methodology}

We propose ObjSplat, a closed-loop framework that integrates Reconstruction, Perception, and Planning into a unified system for high-fidelity active object reconstruction. As illustrated in Fig.~\ref{fig:pipeline}, the system takes RGB-D inputs and progressively updates a Gaussian surfel-based object representation while assessing reconstruction completeness and quality to guide active view planning. ObjSplat consists of three main components: (i) Gaussian surfel-based reconstruction (see \cref{subsec:gs_reconstruction}), which incrementally updates the object model and jointly optimizes geometry and appearance by using surface normals; (ii) Geometry-aware viewpoint evaluation (see \cref{subsec:uncertainty}), a perception module that explicitly quantifies reconstruction quality, identifying under-reconstructed and back-facing regions via occlusion-aware visibility analysis; and (iii) next-best-path planning (see \cref{subsec:nbp}), which utilizes these fine-grained quality metrics to perform multi-step lookahead on a spatial topology, generating globally efficient trajectories that balance information gain and movement cost.

\subsection{Gaussian Surfel-based Reconstruction}
\label{subsec:gs_reconstruction}
\subsubsection{Gaussian Surfels for Object Representation}
\label{subsubsec:representation}
We represent the target object using an unstructured set of 2D Gaussian surfels (GSurfels)~\cite{dai2024high}, where the spatial distribution of each surfel approximates a disk. Each surfel is parameterized by a center $\boldsymbol{\mu} \in \mathbb{R}^3$, a covariance matrix $\boldsymbol{\Sigma}$, an opacity $o \in [0,1]$, and view-dependent color information $\boldsymbol{c} \in \mathbb{R}^{K \times 3}$ utilizing spherical harmonics (SH) coefficients of degree $L_D$, where $K = (L_D+1)^2$. To facilitate optimization, the covariance matrix $\boldsymbol{\Sigma}$ is decomposed into a rotation matrix $\boldsymbol{R}$ (derived from a quaternion $\boldsymbol{q} \in \mathrm{SO}(3)$) and a scaling matrix $\boldsymbol{S} = \text{diag}(s^x, s^y, 0)$, such that $\boldsymbol{\Sigma} = \boldsymbol{R} \boldsymbol{S} \boldsymbol{S}^T \boldsymbol{R}^T$. Notably, the zero $z$-scale constraint enforces a local planar structure, allowing the surface normal $\boldsymbol{n}$ to be directly derived from $\boldsymbol{q}$. This anisotropic formulation enables each surfel to capture local surface geometry more effectively. 

During rendering, GSurfels are transformed into the camera coordinate system via pose $\boldsymbol{T}_c^w \in \mathrm{SE}(3)$ and projected onto the image plane using differentiable splatting~\cite{kerbl20233d}. For a given pixel $\boldsymbol{u}$, intersected surfels are sorted by depth along each viewing ray, and the pixel color is computed through $\alpha$-blending as follows:
\begin{equation}
\label{eq:render_opacity}
    \hat{C} = \sum_{i=1}^n T_i \alpha_i \mathbf{c}_i, \quad \hat{O} = \sum_{i=1}^n T_i \alpha_i, \quad T_i = \prod_{j=1}^{i-1} (1 - \alpha_j),
\end{equation}
where $\alpha_i = o_i \mathcal{N}(\boldsymbol{u}; \boldsymbol{\mu}^{2D}_i, \boldsymbol{\Sigma}^{2D}_i)$ is the opacity modulated by the 2D Gaussian probability, and $\mathbf{c}_i$ denote the SH-encoded color of the $i$-th surfel. The term $T_i$ represents the accumulated transmittance. Similarly, depth $\hat{D}$ and normals $\hat{N}$ are rendered with $\alpha$-blending and normalized to account for partial opacity:
\begin{equation}
\label{eq:render_depth_normal}
    \hat{D} = \frac{1}{1 - T_{n+1}} \sum_{i=1}^n T_i \alpha_i \mathbf{d}_i, \quad \hat{N} = \frac{1}{1 - T_{n+1}} \sum_{i=1}^n T_i \alpha_i \boldsymbol{R}^z_i,
\end{equation}
where $\mathbf{d}_i$ is the unbiased depth computed by intersecting the $i$-th surfel with the viewing ray, and $R^z_i$ represents the surface normal of the $i$-th surfel in the camera frame.

\subsubsection{Geometry–Texture Joint Optimization}
\label{subsubsec:joint_optimization}
To achieve accurate and consistent reconstruction, we jointly optimize geometric and photometric parameters of all GSurfels using a combination of loss terms. Given an input RGB-D frame $I = \{C, D\}$, which includes an RGB image $C \in \mathbb{R}^{H \times W \times 3}$ and its corresponding depth map $D \in \mathbb{R}^{H \times W}$, we additionally extract an object-centric binary mask $M \in \{0, 1\}^{H \times W}$ using pre-trained visual foundation model (VFM)~\cite{ravi2024sam}. We apply a photometric loss $\mathcal{L}_p$ to supervise the color appearance, combining an $L_1$ term and a differentiable structural similarity index measure (D-SSIM) term~\cite{wang2004image}. Meanwhile, depth supervision is enforced via an $L_1$ loss on the predicted depth:
\begin{equation}
\label{eq:loss_rgb}
    \mathcal{L}_{p} = \lambda_1 \left| C - \hat{C} \right| + \lambda_2 \left( 1 - \text{SSIM}(C, \hat{C}) \right), \mathcal{L}_{d} = \left| D - \hat{D} \right|,
\end{equation}
where $\lambda_1=0.8, \lambda_2=0.2$, $\hat{C}$ and $\hat{D}$ are the rendered results. To further enhance surface quality and alignment, we incorporate normal supervision $\mathcal{L}_{n}$ and geometric consistency loss $\mathcal{L}_{c}$ between the rendered normal and the one derived from depth:
\begin{equation}
    \label{eq:loss_depth}
    \quad \mathcal{L}_{n}= \left| N_{\mathrm{obs}}-\hat{N} \right|,  \mathcal{L}_{c}=\left| 1-N_d^{\top} \hat{N} \right|,
\end{equation}
where $N_{\mathrm{obs}}$ and $N_d$ represent surface normals estimated via finite differences from the observed and rendered depth maps, respectively. To prevent semi-transparent blending artifacts and enforce sharp surface boundaries, we regularize the opacity of each surfel to favor either fully opaque or fully transparent states:
\begin{equation}
    \label{eq:loss_opacity}
    \mathcal{L}_{\text {o}}= \text{exp}\left( -(\alpha_i - 0.5)^2/0.05 \right).
\end{equation}

Additionally, we employ a binary cross-entropy (BCE) loss $\mathcal{L}_m$ to supervise the reconstruction of the object mask. The overall objective function combines photometric, depth, normal, opacity, and mask supervision terms:
\begin{equation}
\label{eq:loss_all}
\mathcal{L} = \mathcal{L}_{p} + \lambda_d \mathcal{L}_d + \lambda_n (\mathcal{L}_n + \mathcal{L}_c) + \lambda_m \mathcal{L}_m + \lambda_o \mathcal{L}_o,
\end{equation}
where $\lambda_d=0.8, \lambda_n=\lambda_m=0.1$, and $\lambda_o=0.01$ are weighting coefficients that balance the contributions of each term.

\begin{figure}[t]
    \centering
    \includegraphics[width=0.95\linewidth]{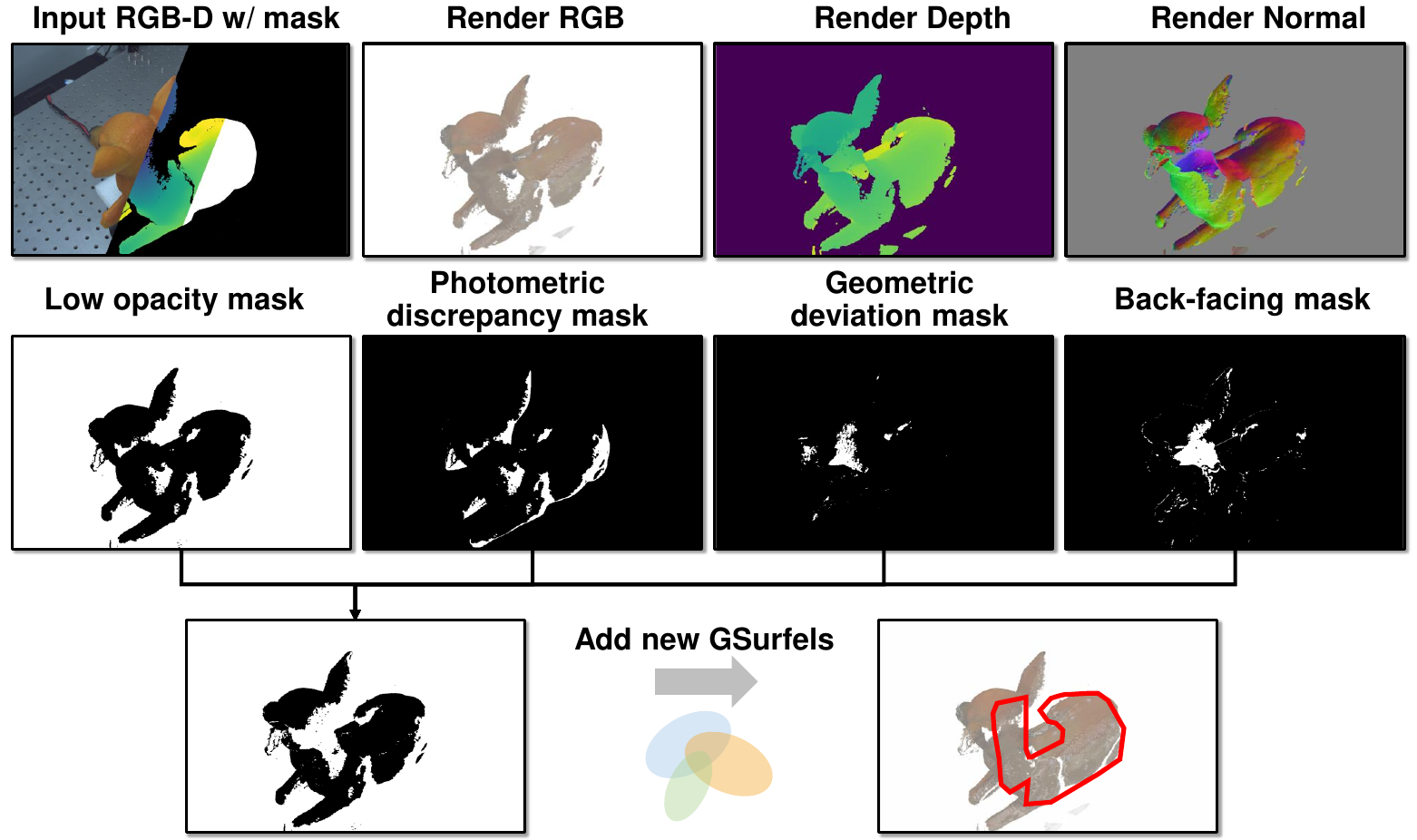}
    \caption{
    \textbf{Identification of insufficiently reconstructed or uncovered regions.} New Gaussian surfels are selectively added to regions exhibiting insufficient opacity, significant photometric discrepancy, geometric deviation, or back-facing surfaces.
    }
    \vspace{-2mm}
    \label{fig:progressive_update}
\end{figure}

\subsubsection{Progressive Update of Gaussian Surfels}
\label{subsubsec:progressive_update}
During progressive object reconstruction through autonomous scanning, each newly captured frame provides additional observations that significantly overlap with previously reconstructed regions. To efficiently fuse new observations while minimizing redundant primitives, it is essential to identify surface areas that are either insufficiently reconstructed or uncovered. New Gaussian surfels are then selectively initialized in these regions to refine both geometry and appearance progressively. We observe that such under-reconstructed regions typically fall into four categories (see Fig.~\ref{fig:progressive_update}): (i) Low opacity ($m_{\mathrm{op}}$): Empty regions where the accumulated rendered opacity $\hat{O}$ falls below a threshold $\tau_{O}=0.5$; (ii) photometric discrepancy ($m_{\mathrm{tex}}$): Pixels where the rendered color deviates from the observed color with a mean squared error ($\text{MSE}$) exceeding $\tau_{C}=0.25$; (iii) geometric deviation ($m_{\mathrm{geo}}$): Regions where the rendered depth lies behind the observed depth, exceeding the mean depth error ($\text{MDE}$) by a factor of $\lambda=2$; (iv) {Back-facing surfaces ($m_{\mathrm{back}}$): Regions where the rendered normal forms an acute angle with the camera optical axis $\boldsymbol{z}$ (i.e., $\hat{N} \cdot \boldsymbol{z} > 0$). This condition implies the camera is observing the "back" of an existing surface, yet the sensor detects valid geometry, indicating a need for refinement (e.g., for thin structures). We define the criterion-specific masks as follows:
\begin{subequations}
\begin{align}
m_{\mathrm{op}}(\boldsymbol{u})   &:= [\,\hat{O}(\boldsymbol{u}) < \tau_{O}\,], \\
m_{\mathrm{tex}}(\boldsymbol{u})  &:= [\,\|\hat{C}(\boldsymbol{u}) - C(\boldsymbol{u})\|_{2}^{2} > \tau_{C}\,], \\
m_{\mathrm{geo}}(\boldsymbol{u})  &:= [\,\hat{D}(\boldsymbol{u}) - D(\boldsymbol{u}) > \lambda \cdot \text{MDE}\,], \\
\label{eq:back_facing_mask}
m_{\mathrm{back}}(\boldsymbol{u}) &:= [\,\hat{N}(\boldsymbol{u}) \cdot \boldsymbol{z} > 0\,], \qquad \boldsymbol{u}\in\Omega.
\end{align}
\end{subequations}
Accordingly, the unified update mask $M$ is the logical disjunction of these conditions: $M = \{ \boldsymbol{u} \in \Omega \mid m_{\mathrm{op}} \lor m_{\mathrm{tex}} \lor m_{\mathrm{geo}} \lor m_{\mathrm{back}} \}$, where $\Omega$ denotes the set of pixels with valid sensor depth.

Formally, the update mask $M$ highlights candidate areas for new surfel insertion, allowing the system to progressively refine geometry and texture while avoiding redundant updates in well-reconstructed regions. For all pixels $\boldsymbol{u} \in M$, we initialize new Gaussian surfels to fill the information gap. The center position $\boldsymbol{\mu}_{\text{new}}$ is recovered via back-projection using the current pose $\boldsymbol{T}_c^w$:
\begin{equation}
    \boldsymbol{\mu}_{\text{new}} = \boldsymbol{T}_c^w \cdot \left( D(\boldsymbol{u}) \Pi^{-1}(\boldsymbol{u}) \right),
\end{equation}
where $\Pi^{-1}$ is the inverse projection operator. The color $\boldsymbol{c}$ is assigned from the image observation $C(\boldsymbol{u})$, and the orientation quaternion $\boldsymbol{q}$ is aligned with the sensor-derived surface normal to get well geometric initialization. Finally, the scaling vector $\boldsymbol{s}$ is initialized adaptively based on the local point density, approximated by the mean distance to the $k$-nearest neighbors in the back-projected point cloud, ensuring that newly added surfels maintain a consistent spatial distribution suitable for optimization.

\subsection{Geometry-Aware Viewpoint Evaluation}
\label{subsec:uncertainty}

\subsubsection{Occlusion-Aware Co-visibility Analysis}
\label{subsubsec:covisibility}
Accurate co-visibility estimation is a prerequisite for both reliable historical keyframe selection in optimization (ref. \ref{subsubsec:view_selection}) and informative viewpoint evaluation (ref. \ref{subsubsec:uncertainty}). Traditional approaches in scene-level mapping typically rely on depth-based heuristics~\cite{li2025activesplat}, which simply count the number of reprojection points falling within the keyframe's view frustum or determine visibility by rendering cumulative opacity~\cite{matsuki2024gaussian}. While effective for large-scale scene mapping, these methods are often insufficient for object-level reconstruction, where the camera observes the target from outside, leading to frequent self-occlusions and non-closed surfaces. As illustrated in Fig.~\ref{fig:covisible}, covisibility metrics that ignore occlusion tend to overestimate shared visibility, resulting in false correspondences and unreliable covisibility estimation.

\begin{figure}[t]
    \centering
    \includegraphics[width=0.95\linewidth]{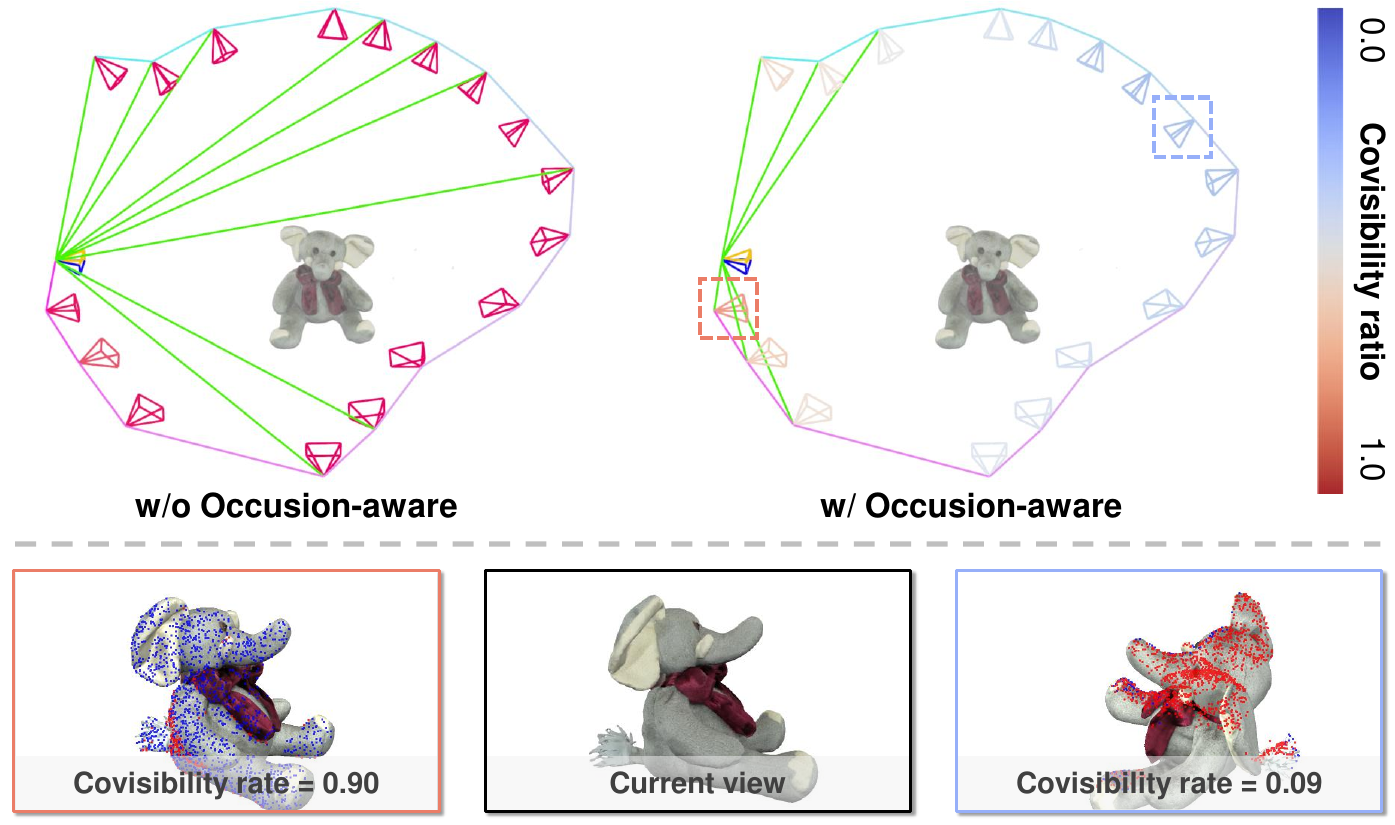}
    \caption{
    \textbf{Co-visible views comparison.} While conventional frustum-based metrics (\textbf{Left}) fail to account for self-occlusion, leading to false correspondences, our rendering-based check (\textbf{Right}) explicitly detects occlusion and back-facing surfaces, thus correctly identifying \textcolor{MyDarkBlue}{valid} and \textcolor{MyRed}{invalid} co-visible points, ensuring valid geometric constraints and confidence updating.
    }
    \vspace{-2mm}
    \label{fig:covisible}
\end{figure}

To overcome these limitations, we introduce an \textit{occlusion-aware covisibility metric} that explicitly models geometric visibility between views using differentiable rendering. Specifically, given a source view $I_i = \{D_i, T_i^w\}$ and a target view $I_j = \{T_j^w\}$, we randomly sample a subset of valid pixels $\mathcal{S}_i \subset \Omega_i$ (default $|\mathcal{S}_i|=1600$). Each pixel $\boldsymbol{u}_i \in \mathcal{S}_i$ is back-projected to world coordinates using $D_i$ and $T_i^w$, then transformed to the coordinate frame of $I_j$ and reprojected as:
\begin{equation}
\label{eq:proj}
\boldsymbol{u}_j = \Pi\!\left((T_j^w)^{-1} T_i^w \Pi^{-1}(\boldsymbol{u}_i)\right),
\end{equation}
where $\Pi$ and $\Pi^{-1}$ denote projection and back-projection functions parameterized by camera intrinsics. We then render the candidate view $I_j$ from the current GSurfels model to obtain its depth $\hat{D}_j$ and normal map $\hat{N}_j$. For each valid correspondence $(\boldsymbol{u}_i,\boldsymbol{u}_j)$, let $D^{\mathrm{proj}}_{i\!\to\! j}(\boldsymbol{u}_j)$ be the depth of the reprojected 3D point in $I_j$. A correspondence is marked as \emph{invalid} if the reprojected point is occluded by the rendered surface or falls on a back-facing region (ref Eq.~\ref{eq:back_facing_mask}). Formally, the invalid covisibility mask is defined as:
\begin{equation}
\label{eq:inv}
m_{\mathrm{inv}}(\boldsymbol{u}_j)=
\big[\, D^{\mathrm{proj}}_{i\!\to\! j}(\boldsymbol{u}_j)-\hat{D}_j(\boldsymbol{u}_j)>\tau_d \,\big]
\,\lor\,
m_{\mathrm{back}}(\boldsymbol{u}_j),
\end{equation}
where $\tau_d$ is a depth tolerance threshold. The covisibility ratio $\rho_{i,j}$ is further defined as the proportion of sampled pixels $\mathcal{S}_i$ that maintain valid covisibility in the target view $I_j$:
\begin{equation}
\label{eq:covis-rate}
\rho_{i,j}
= \frac{|\mathcal{S}_{i\!\to\! j}| - \sum_{\boldsymbol{u}_j\in \mathcal{S}_{i\!\to\! j}} 
\mathbb{I}\!\left[m_{\mathrm{inv}}(\boldsymbol{u}_j)\right]}{|\mathcal{S}_i|}
= \frac{|\Omega_{i,j}|}{|\mathcal{S}_i|},
\end{equation}
where $\mathcal{S}_{i\!\to\! j}\!\subseteq\!\mathcal{S}_i$ represents the subset of points whose reprojections fall within the field of view of $I_j$, and $\Omega_{i,j}$ denotes the subset of $\mathcal{S}_{i\!\to\! j}$ that remains valid after excluding occluded and back-facing points. This metric provides a robust measure of geometric overlap, essential for maintaining consistency. If no valid projection exists ($|\mathcal{S}_{i\!\to\! j}|=0$), we set $\rho_{i,j}=0$. 

\subsubsection{Global–Local View Selection}
\label{subsubsec:view_selection}
During online reconstruction, Gaussian surfel attributes are updated through differentiable rendering and gradient-based optimization. Unlike the offline refinement stage, jointly optimizing all historical frames is computationally inefficient and unnecessary for incremental reconstruction. Instead, we employ a global–local view selection strategy inspired by GS-based SLAM~\cite{matsuki2024gaussian, wu2025monocular} to balance local adaptability with long-term global consistency.

\texttt{Local views}: To refine newly observed surface regions through multi-view constraints, we maintain a local window $\mathcal{W}_{\text{local}}$. Given the current view $I_t$, we select the top-$k$ views with the highest covisibility ratio $\rho_{t,j}$ (Eq.~\ref{eq:covis-rate}). These views jointly observe the newly captured regions, providing dense geometric and photometric constraints for incremental optimization of the corresponding GSurfels.

\texttt{Global views}: To mitigate the overfitting of newly observed regions caused by relying solely on local views and to ensure global consistency, we additionally maintain a sparse set of global views $\mathcal{W}_\text{global}$, sampled from earlier keyframes outside the local window. To ensure temporal balance, we follow the reweighting scheme in~\cite{wu2025monocular}, increasing the sampling probability of early or under-optimized keyframes.

This hybrid selection strategy prioritizes covisible keyframes for accurate local updates while preserving long-range consistency across the reconstruction.

\subsubsection{Geometry-Aware Uncertainty Qualification}
\label{subsubsec:uncertainty}
Accurate online evaluation of reconstruction quality and completeness is crucial for active reconstruction, as it enables the robot to identify under-reconstructed regions and plan informative subsequent viewpoints. However, conventional uncertainty metrics~\cite{jiang2024fisherrf, li2025activesplat, jin2025activegs} often overlook self-occlusion and back-face visibility, leading to unreliable viewpoint evaluation, particularly in object-level reconstruction, where the camera observes the target from an external perspective and the surface is often open or partially occluded. To address this, we propose a \textit{geometry-aware uncertainty map} that jointly encodes surfel confidence, back-face observation, and visibility-based completeness. This map provides a dense, view-dependent quality estimate for any candidate viewpoint, serving as a guiding signal for subsequent path planning. As illustrated in Fig.~\ref{fig:pipeline} and Fig.~\ref{fig:qualitative_comparison_uncer}, it consists of three components: (1) surfel-wise confidence modeling, (2) back-face observation detection, and (3) visibility-based completeness estimation.

\begin{figure}[t]
    \centering
    \includegraphics[width=0.99\linewidth]{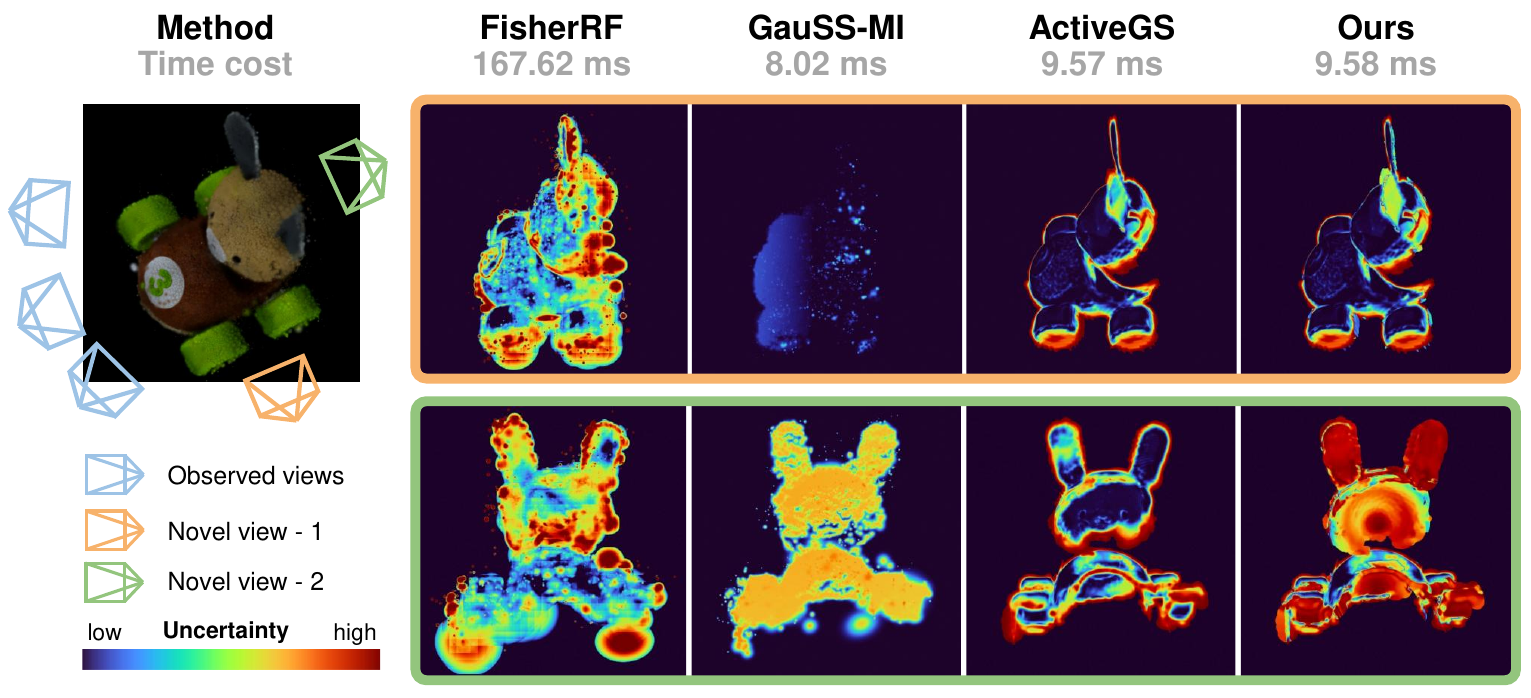}
    \caption{Uncertainty qualification comparison on the \texttt{Bunny Racer} object. Our method enables fine-grained uncertainty evaluation with millisecond-level latency, effectively addressing complex geometric challenges such as self-occlusions and thin structures.}
    \label{fig:qualitative_comparison_uncer}
    \vspace{-2mm}
\end{figure}

\paragraph{Surfel-wise Confidence Modeling} We assign a confidence attribute $\kappa_i \in [0,1]$ to each Gaussian surfel $g_i$ to quantify the reliability of its historical observations. Extending the formulation in~\cite{jin2025activegs} to the object-centric setting, we explicitly incorporate (i) front-view weighting to prioritize near-orthogonal observations and (ii) occlusion-aware validity checks to discard back-facing or occluded surfels. Upon acquiring a new view $I_t$, we identify the set of effectively observed surfels $\mathcal{G}^{\text{valid}}(I_t)$ using the validity check in Eq.~\ref{eq:inv}. For each $g_i \in \mathcal{G}^{\text{valid}}(I_t)$, we update its confidence by aggregating contributions from its set of valid historical views $S^{\text{valid}}(g_i)$ (including the current view $I_t$). Specifically, for each view $j \in S^{\text{valid}}(g_i)$, we compute a geometric weight $w_{ij} = w^{\text{dist}}_{ij} w^{\text{front}}_{ij}$ to penalize distant or oblique observations:
\begin{subequations}
\begin{align}
    w^{\text{dist}}_{ij} &= \max\!\left(0, 1 - \frac{d_{ij}}{d_{\max}}\right), \\
    w^{\text{front}}_{ij} &= \sigma\!\left(\frac{\mathbf{n}_i^{\top} \mathbf{v}_{ij} - c_0}{\tau}\right),
\end{align}
\end{subequations}
where $\mathbf{v}_{ij}$ denotes the normalized viewing direction from the surfel center $\boldsymbol{\mu}_i$ to the camera center $\mathbf{x}^p_j$, $d_{ij}$ is the distance, and the $\sigma(\cdot)$ is a sigmoid function prioritizing near-frontal views, configured with threshold $c_0=0.5$ (approx. $60^\circ$) and temperature $\tau=0.1$. Furthermore, to encourage multi-view coverage, we incorporate angular diversity $\beta_i = 1 - \|\boldsymbol{\mu}_i\|$ based on the mean viewing direction of $S^{\text{valid}}(g_i)$. The final confidence $\kappa_i$ is derived by combining the accumulated geometric weights with the angular diversity factor:
\begin{equation}
\label{eq:confidence}
    \gamma_i = \sum_{j\in S^{\text{valid}}(g_i)} w_{ij}\, \max(0, \mathbf{n}_i^{\top}\mathbf{v}_{ij}), \quad \kappa_i \leftarrow \gamma_i \exp(\beta_i).
\end{equation}

This update rule ensures that high confidence is assigned only when surfels are observed from diverse, non-occluded, and near-orthogonal viewpoints, whereas those seen from sparse or oblique views receive lower scores.

\paragraph{Back-Face Observation} We detect back-facing regions where the rendered normal $\hat{N}$ forms an acute angle with the camera's optical axis $\mathbf z$. These areas are marked as potentially incomplete, as they are viewed from the "wrong" side. This encourages exploration from opposing directions. The back-face score for a pixel $u$ is defined as:
\begin{equation}
\label{eq:backface_score}
    B(\boldsymbol{u}) = \max(0, \hat{N}(\boldsymbol{u})^\top \boldsymbol{z}).
\end{equation}

\paragraph{Visibility-based Completeness} We use the rendered opacity $\hat{O}(u)$ from current GSurfels as a direct measure of geometric coverage. Regions with low opacity indicate insufficient surfel density or unobserved space, providing a dense metric for reconstruction completeness. We define the visibility-based uncertainty score as:
\begin{equation}
\label{eq:visibility_uncertainty}
V(u) = 1 - \hat{O}(\boldsymbol{u}).
\end{equation}

Combining these components, the overall geometry-aware uncertainty map $\hat U(\boldsymbol{u})$ for a candidate view is formulated as a weighted sum:
\begin{equation}
\label{eq:uncertainty_map}
\hat{U}(\boldsymbol{u}) = \lambda_k (1-K(\boldsymbol{u})) + \lambda_b B(\boldsymbol{u}) + \lambda_v V(\boldsymbol{u}),
\end{equation}
where $K(\boldsymbol{u})$ is the rendered confidence map derived from surfel attributes $\kappa_i$. The coefficients $(\lambda_k, \lambda_b, \lambda_v)$ dynamically balance the relative importance of confidence, back-face, and visibility terms throughout the reconstruction process, as detailed in \cref{subsubsec:dynamic_ucertainty_weighting}. Benefiting from the CUDA-accelerated rasterization in Gaussian splatting, our system renders this map in about 2 ms, enabling real-time evaluation for path planning.

\subsection{Next-Best-Path Planning}
\label{subsec:nbp}
To achieve both efficient and thorough reconstruction, the robot must plan a globally-aware scanning trajectory that optimizes the trade-off between exploration gain and movement cost. Moving beyond the greedy, single-step nature of traditional NBV planning, we propose a Next-Best-Path (NBP) planning strategy that performs multi-step lookahead on dynamically constructed spatial topology, jointly considering the accumulated information gain—derived from our geometry-aware uncertainty metric (Eq.~\ref{eq:uncertainty_map})—and traversal cost. The planning process consists of two main stages: candidate viewpoint generation and path planning.

\subsubsection{Candidate Viewpoint Generation}
\label{subsubsec:candidate_sampling}
We generate candidate viewpoints on a hemispherical shell defined around the object's current bounding box, which is updated online as the Gaussian surfel model expands. This ensures the entire object remains within the camera's field of view. Specifically, we define a sphere with a radius equal to the sum of the box's half-diagonal and an optimal standoff distance (related to focal length). To ensure uniform coverage, we distribute candidate positions on the sphere using a Vogel spiral pattern~\cite{vogel1979better}. To prevent redundant exploration, previously visited viewpoints are pruned from the candidate set. Each candidate viewpoint is oriented to face the center of the bounding box, ensuring object-centric observation. This uniform sampling strategy generates a sparse yet comprehensive set of candidates that adaptively adjusts to the object’s perceived size and position.

\subsubsection{Next-Best-Path Strategy}
\label{subsubsec:nbp_strategy}
With a set of candidate viewpoints $\mathcal{V}_{\text{cand}}$ and their corresponding uncertainty scores $\{U(v) \mid v \in \mathcal{V}_{\text{cand}}$, our goal is to determine a trajectory that maximizes total information gain while minimizing travel distance. We formulate this as a variant of the prize-collecting traveling salesman problem (PC-TSP)~\cite{balas1989prize}. First, we identify a long-term goal $v_g \in \mathcal{V}_{\text{cand}}$, defined as the viewpoint with the highest uncertainty score. The problem then reduces to finding an optimal path from the current pose $v_t$ to $v_g$ that collects the most "prizes" (information gain) from intermediate viewpoints at a minimal travel cost.

We solve this by constructing a spatial topology graph $G=(V, E)$, where the node set $V = \mathcal{V}_{\text{cand}} \cup \{v_t\}$. Edges are formed by connecting each viewpoint to its $k\!=\!10$ nearest neighbors ($k$-NN) in Euclidean space. The current view $v_t$ is also integrated into the graph by connecting it to its own $k$ nearest candidate viewpoints. The graph structure provides a sparse yet effective representation of the feasible motion pathways. To guide the path search, we assign an uncertainty-aware weight to each edge $(v_i, v_j) \in E$ that balances the motion distance with the potential information gain:
\begin{equation}
\label{eq:edge_weight}
w(v_i, v_j) = \frac{d_{ij}}{\alpha + \beta \left( U(v_i) + U(v_j) \right)},
\end{equation}
where $d_{ij}$ is the Euclidean distance between viewpoints $v_i$ and $v_j$, and we set $\alpha\!=\!0.1$ and $\beta\!=\!0.5$ to prevent division by zero and scaling the reward. This formulation ensures that edges connecting high-uncertainty viewpoints have lower weights, making them more attractive to a shortest-path algorithm. In here, finding the single optimal path is a multi-objective optimization problem. Instead of greedily selecting a single shortest path, we search for a set of candidate paths. We first find the top-$M$ shortest simple paths~\cite{yen1971finding} from $v_t$ to $v_g$ on the weighted graph $G$. Each candidate path $\pi = (v_t, n_1,\dots,n_L{=}v_g)$ is then evaluated by explicitly trading off its total collected reward against its path length:
\begin{equation}
\label{eq:path_score}
J(\pi) = \lambda \sum_{v \in \pi} \bar{U}(v) - (1-\lambda) \sum_{k=0}^{L-1} \bar{d}(n_k, n_{k+1}),
\end{equation}
where $\bar{(\cdot)}$ is the normalization operator, and we use $\lambda\!=\!0.5$ to balance the accumulated information gain and the movement cost. The path $\pi^\star$ with the maximum score is selected as the next-best-path $\pi^* = \arg\max_\pi J(\pi)$ to execute. This strategy allows the system to perform multi-step lookahead, escaping local optima often encountered in greedy NBV planning, resulting in smoother and more globally efficient trajectories.

\begin{algorithm}[t]
    \small
    \caption{Next-Best-Path (NBP) Planning}
    \label{alg::nbp}
    \begin{algorithmic}[1]
        \STATE {\bfseries Input:} Current pose $v_t$, Constructed Model $\mathcal{G}_t$, Uncertainty map renderer $\hat{U}$
        \STATE {\bfseries Output:} An optimal path of viewpoints $\pi^\star$
        
        \STATE $\mathcal{V}_{\text{cand}} \gets \text{GenerateCandidates}(\mathcal{G}_t)$ \hfill\(\triangleright\) \cref{subsubsec:candidate_sampling}
        \STATE $\mathcal{V}_{\text{cand}} \gets \text{PruneVisited}(\mathcal{V}_{\text{cand}})$
        \STATE $v_g \gets \arg\max_{v \in \mathcal{V}_{\text{cand}}} \hat{U}(v)$ \hfill\(\triangleright\) Identify Long-term goal
        \STATE $G \gets \text{BuildKNNGraph}(\mathcal{V}_{\text{cand}} \cup \{v_t\})$
        \STATE Assign edge weights in $G$ using Eq.~\ref{eq:edge_weight}
        \STATE $\Pi \gets \text{FindTopKShortestPaths}(G, v_t, v_g)$
        \STATE $\pi^\star \gets \arg\max_{\pi \in \Pi} J(\pi)$ \hfill\(\triangleright\) Select Optimal Path
        
        \STATE \bf{return} $\pi^\star$
    \end{algorithmic}
\end{algorithm}

As summarized in Algorithm~\ref{alg::nbp}, the NBP planner generates an optimal path $\pi^\star$ from the current pose to a sub-goal $v_g$. The robot sequentially visits viewpoints in the optimal path $\pi^\star$, capturing new sensor data to incrementally update the object model at each step. Upon reaching the sub-goal $v_g$, the system re-evaluates the global uncertainty and triggers a new NBP planning cycle. This iterative replanning strategy ensures continuous adaptation to the evolving reconstruction while reducing computational overhead compared to per-step planning. Additionally, the system naturally supports an auto-stop mode: the active scanning process autonomously terminates once the average reconstruction quality uncertainty $\hat{U}_q(\boldsymbol{u}) = \lambda_b B(\boldsymbol{u}) + \lambda_v V(\boldsymbol{u})$ falls below a predefined threshold $\tau_{\text{stop}}\!=\!0.05$).

\subsubsection{Dynamic Uncertainty Weighting}
\label{subsubsec:dynamic_ucertainty_weighting}
To balance exploration and exploitation effectively throughout the reconstruction process, we employ a dynamic weighting strategy for the components of the uncertainty map from Eq.~\ref{eq:uncertainty_map}.
In the early stages of reconstruction, when object information is Insufficient, the system prioritizes exploration ($\lambda_k=\lambda_b=\lambda_v=1$) to rapidly cover the object's surface, focusing on low-confidence, back-facing, and poor visibility (low-opacity) regions. As the reconstruction progresses and more of the object becomes observed, the focus shifts to exploitation---refining the quality of the reconstructed model. Once visibility-based uncertainty stabilizes, we de-prioritize the visibility term by setting $\lambda_v=0$, concentrating planning efforts on resolving subtle defects like back-facing surfaces and low-confidence areas, thereby ensuring high-fidelity final quality.

\subsection{Implementation Details}
\label{subsec:details}
To expand the robot's observation space, our experimental platform comprises a robotic arm equipped with an eye-in-hand RGB-D camera and a motorized turntable positioned 0.7\,m in front of the robot base (see Fig.~\ref{fig:experimental_platforms}) to handle small to medium-sized unknown objects. The object is placed on the turntable, which is treated as the origin of the world coordinate system. By pre-calibrating the eye-in-hand transformation $\boldsymbol{T}^b_c$ (camera to robot base) and the turntable-to-base transform $\boldsymbol{T}^b_t$, we derive the camera pose in the object frame as $\boldsymbol{T}_c = (\boldsymbol{T}^b_t)^{-1}\boldsymbol{T}^b_c$. Consequently, all view planning is conducted within this object-centric frame. To handle kinematic constraints, we implement a coordinated motion strategy: if a target viewpoint is unreachable by the arm, the turntable rotates to bring it into the feasible workspace, effectively extending the system's operational range.

\subsubsection{Object-Centric Pose Tracking}
\label{subsubsec:object_centric_tracking}
Accurate and robust camera pose tracking is critical for coherent fusion. In real-world scenarios, inherent motion noise in the robotic arm and minor calibration inaccuracies can lead to pose drift over time. We mitigate this via a robust point-cloud registration-based object-centric tracking pipeline. For each incoming RGB-D frame, we first employ the segment model~\cite{ravi2024sam} to extract the object of interest from the background, yielding a clean, object-level point cloud. Subsequently, we perform pairwise point-to-point registration between the current cloud and the accumulated global map from the previous keyframe. This step effectively compensates for slight pose inaccuracies from the robot's odometry, ensuring globally consistent alignment.

\subsubsection{Online Mapping and Offline Refinement}
Leveraging the consistent parameter space of Gaussian splatting-based approaches, we decompose the reconstruction process into two stages: \textit{online mapping} and \textit{offline refinement}. This allows for post-processing of all observations collected during the active scanning, thereby enhancing reconstruction fidelity. In the \textit{online mapping} stage, efficiency is prioritized for real-time performance. We simplify the GSurfel representation by setting the SH degree to $L_D\!=\!0$, effectively reducing memory usage and computational overhead. Each new frame undergoes 10 optimization iterations, while the local optimization window consists of the current view and its top-$k=9$ most covisible keyframes. Once data acquisition is complete, the \textit{offline refinement} stage focuses on improving global consistency and visual quality. We conduct 7{,}000 optimization iterations while restoring higher-order SH coefficients ($L_D\!=\!3$) to achieve photorealistic rendering. Both camera poses and GSurfel parameters are jointly optimized. To mitigate photometric inconsistencies caused by illumination variation during scanning, we additionally optimize per-frame exposure compensation parameters, yielding globally consistent color appearance~\cite{zhang2025hi}. Finally, to generate physically consistent assets for downstream tasks, we extract surface meshes via TSDF fusion or Poisson reconstruction~\cite{kazhdan2013screened} using depth and normal maps rendered from the refined model.

%% file: src/4_experiments.tex
\section{Experiments}
\label{sec:experiments}
\begin{figure}[t]
    \centering
    \includegraphics[width=1.0\linewidth]{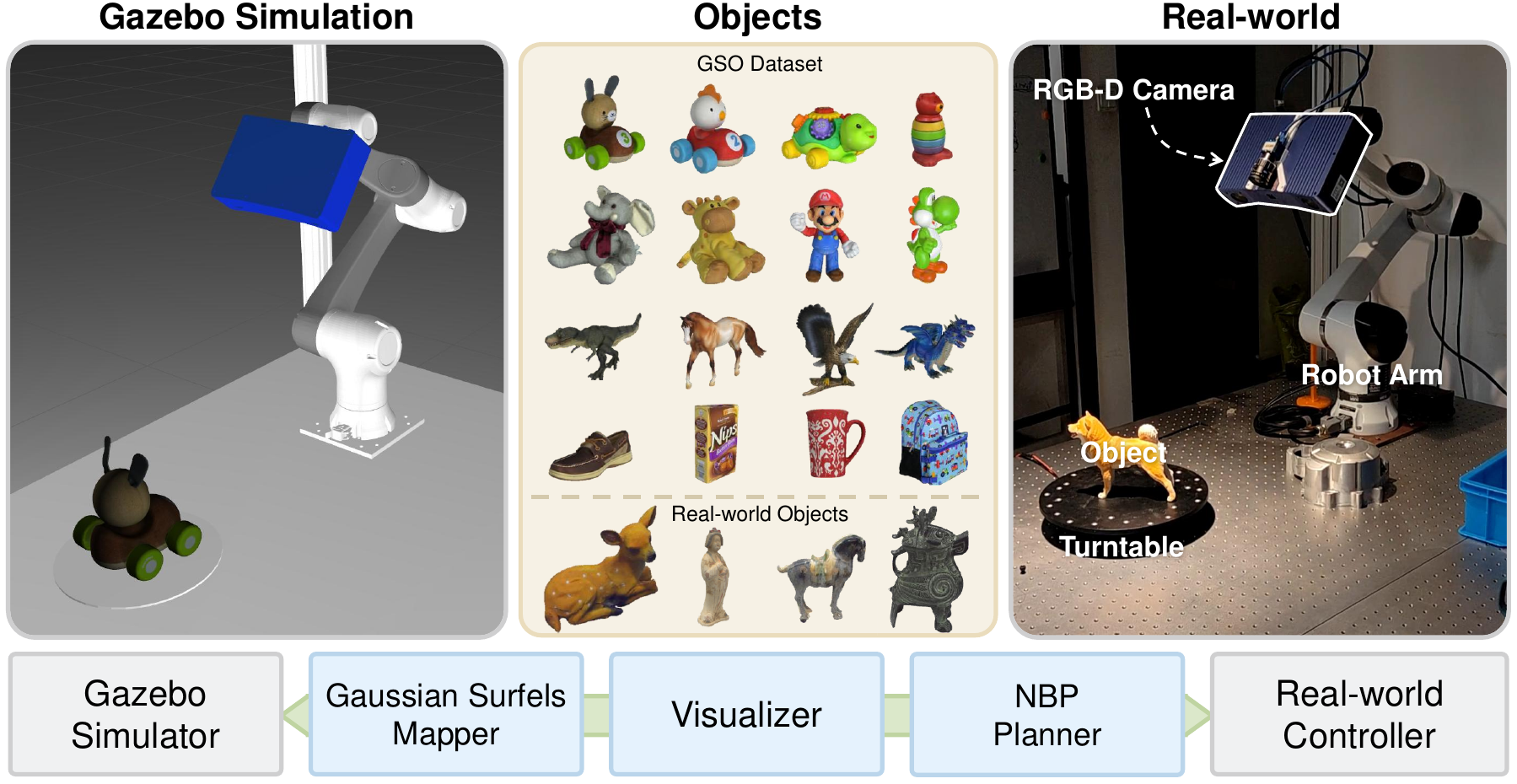}
    \caption{\textbf{Experimental Platforms.} (\textbf{Left}) The \textit{Gazebo} simulation environment provides realistic rendering and physics-based robot models for quantitative evaluation. (\textbf{Right}) The real-world platform comprises an Elfin-05 robotic arm equipped with a calibrated industrial 3D camera. The system architecture is built on modular ROS nodes for Gaussian mapping, NBP planning, and visualization across both domains.}
    \label{fig:experimental_platforms}
\end{figure}

To comprehensively evaluate ObjSplat, we conduct extensive experiments in both high-fidelity simulation and real-world scenarios. We aim to validate the system's reconstruction quality, exploration efficiency, and robustness against diverse geometric and textural complexities. All experiments are performed on a desktop PC equipped with an Intel Core i9-13900K CPU and a single NVIDIA RTX 4090 GPU, running Ubuntu 20.04 LTS with ROS Noetic for inter-module communication. The experimental setup is depicted in Fig.~\ref{fig:experimental_platforms}.

\subsection{Experimental Setup}
\label{subsec:setup}
\subsubsection{Baselines}
\label{subsubsec:baselines}
We compare ObjSplat against a comprehensive set of state-of-the-art (SOTA) active object reconstruction methods, spanning geometry-based planning to information-theoretic radiance field exploration: (i) SEE~\cite{border2024surface}: A frontier-based method that operates directly on point clouds to explore visible surface edges, serving as a strong baseline for geometric completeness; (ii) PB-NBV~\cite{jia2025pb}: A projection-based greedy NBV planner that maximizes voxel-based information gain, representing the single-step planning paradigm; (iii) MA-SCVP~\cite{pan2024integrating}: A point-cloud-based one-shot view planning framework that solves a set covering optimization problem, representing global planning paradigm within geometric reconstruction approaches; (iv) NeRF-PRV~\cite{pan2024many}: A NeRF-based one-shot view planner that utilizes a neural network to predict the required number of views based on object complexity; (v) FisherRF~\cite{jiang2024fisherrf}: An information-theoretic approach that leverages the Fisher Information to quantify the parameter uncertainty of the underlying radiance field, guiding view selection to maximize expected information gain; (vi) GauSS-MI~\cite{xie2025gauss}: A recent information-theoretic approach that leverages the Gaussian splatting Shannon mutual information to quantify the expected reduction in visual uncertainty.

\subsubsection{Datasets and Scenarios}
\label{subsubsec:datasets}

\paragraph{Simulation}
We utilize 16 diverse models from the Google Scanned Objects (GSO) dataset~\cite{downs2022google}, which offers high-quality ground-truth models, selected for their challenging attributes such as thin structures, hollow geometries, and fine-grained textures (see Fig.\ref{fig:experimental_platforms}). Each object is normalized to fit within a bounding box diagonal of [0.1, 0.35]~m. Following prior work~\cite{pan2024integrating}, a simulated RGB-D camera with a resolution of $1280 \times 720$ and a depth sensing range of [0.1, 3.0]\,m is mounted on the robot end-effector. All methods are initialized from an identical pose to ensure fair comparison.

\paragraph{Real-world}
For hardware deployment, the system is executed on an Elfin-05 robotic arm equipped with a calibrated industrial vision system, which consists of an Ainstec fringe-projection structured-light 3D camera for precise depth acquisition and a Galaxy industrial camera for synchronized RGB imaging at a resolution of $1920 \times 1200$. Using this platform, we select four challenging objects to assess robustness: (1) a \texttt{Sika Deer} sculpture, featuring complex geometry and fine details; (2) a \texttt{Fu Hao Owl Zun} replica, characterized by complex geometric topology yet exhibiting a relatively uniform texture; and (3) \texttt{Sancai Horse} and \texttt{Pottery Figure} replicas, both distinguished by their intricate patterns and rich, high-frequency textural details.

\subsection{Evaluation Metircs}
\label{subsubsec:metrics}
Following \cite{ye2024pvp} and \cite{pan2024integrating}, we evaluate both reconstruction quality and process efficiency. All quality metrics are computed within object masks to avoid background influence.

\subsubsection{Reconstruction Quality}
\label{subsubsec:metric_quality}
For appearance, we report the peak signal-to-noise ratio (PSNR), structural similarity index (SSIM), and learned perceptual image patch similarity (LPIPS)~\cite{kerbl20233d} on rendered images from both training and novel views (sampled from unvisited candidate views). For geometry, we first assess multi-view shape accuracy by reporting the depth L1 error (cm) on rendered depth maps. We then extract a mesh from the reconstructed Gaussian model, align it to the ground-truth mesh using ICP, and uniformly sample 200k points from both meshes. Geometric fidelity is quantified using the Chamfer Distance (CD) and F-Score@0.005m~\cite{ye2024pvp}. CD measures the average bidirectional nearest-neighbor distance, capturing fine geometric deviations, while F-Score reflects both local precision and surface completeness (recall).

\subsubsection{Completeness and Efficiency}
\label{subsubsec:metric_comp_efficiency}
To evaluate autonomous reconstruction performance, we use: (1) Completion ratio (CR, $\%$) and completion (cm), quantifying the proportion of the target scene successfully scanned; (2) Movement cost (m), the total distance between all viewpoints; (3) Required views, the number of viewpoints; and (4) Time cost (s), including both the entire online autonomous scanning/reconstruction time and the offline refinement time.

\subsection{Reconstruction Performance Analysis}
\label{subsec:recon_performance}

\subsubsection{Qualitative Analysis of Reconstruction Quality}
\label{subsubsec:qualitative_analysis_quality}
\begin{figure*}[t]
    \centering
    \includegraphics[width=0.95\linewidth]{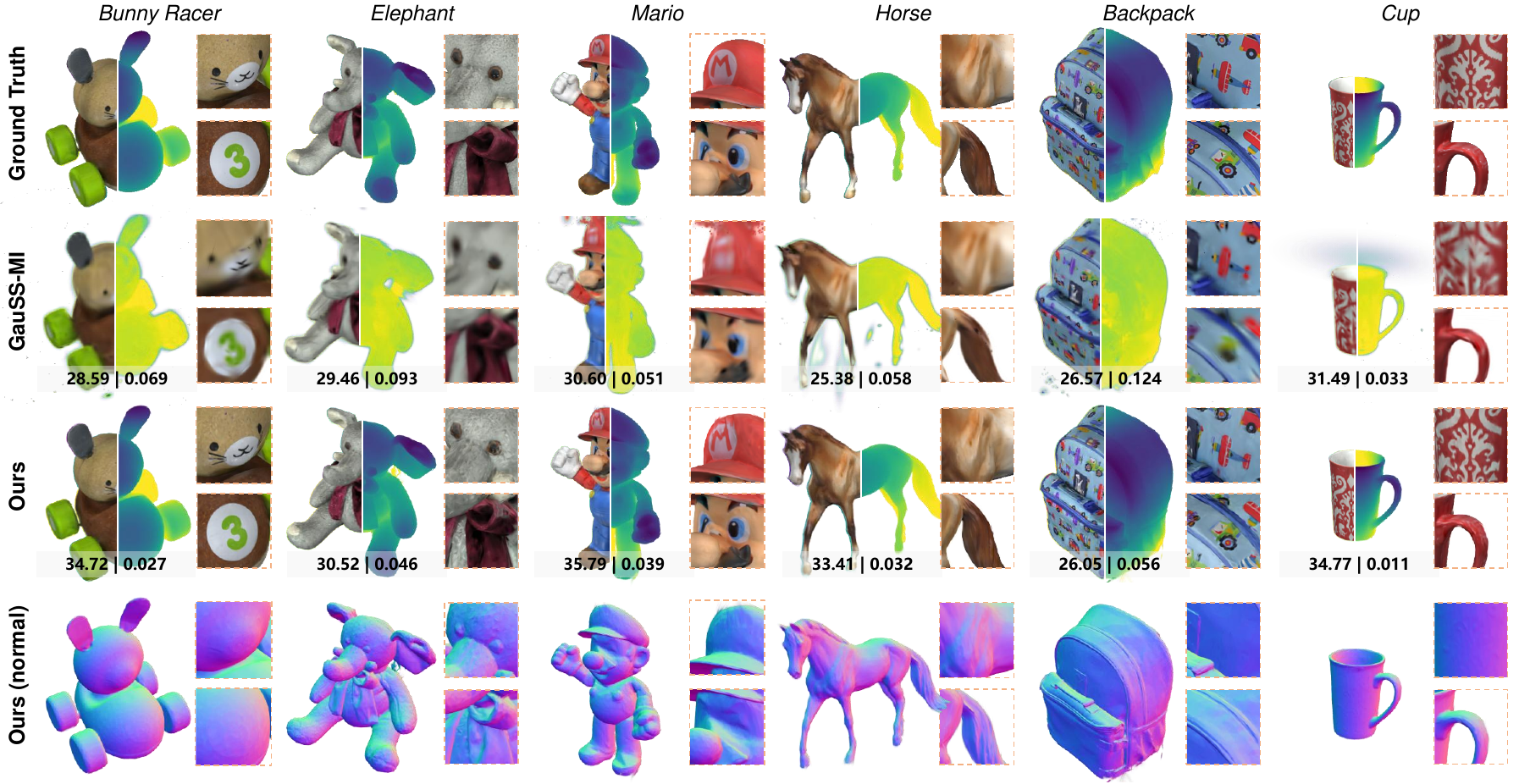}
    \caption{Qualitative comparison of reconstruction results on six representative objects from the GSO dataset. ObjSplat recovers sharp textures and clean geometry, whereas baselines suffer from blurriness and floating artifacts due to overfitting to sparse views. The bottom row of each picture shows the average novel PSNR (dB) and LPIPS.}
    \label{fig:qualitative_comparison_recon_quality}
\end{figure*}

Fig.~\ref{fig:qualitative_comparison_recon_quality} presents a qualitative comparison of six representative objects. All methods underwent a standardized offline refinement phase consisting of 7,000 iterations for fair comparison. Baselines (GauSS-MI~\cite{xie2025gauss} and FisherRF~\cite{jiang2024fisherrf}) use their default color refinement scheme from MonoGS~\cite{matsuki2024gaussian}.

GauSS-MI, relying on vanilla 3DGS, exhibits noticeable degradation under sparse observations. Lacking explicit multi-view geometric constraints, it tends to produce blurry textures and significant floating artifacts around the object, as unconstrained Gaussians overfit to sparse input views. In contrast, ObjSplat reconstructs complete, photorealistic models with sharp details. While PSNR values are comparable in certain cases, our method achieves significant improvements in perceptual metrics (i.e., LPIPS), reflecting superior visual fidelity, particularly on objects with fine-grained patterns as the \texttt{Backpack} and \texttt{Cup}. We attribute this performance to the synergy between our active planning strategy and robust representation: (1) Informative Data Acquisition: Our geometry-aware uncertainty module targets geometric defects (e.g., back-facing surfaces, low opacity) and under-observed regions. This guides the robot to acquire high-information observations that efficiently resolve geometric ambiguities, ensuring comprehensive surface coverage within a limited view budget; and (2) Robust Regularization: The adoption of Gaussian surfels, coupled with geometric consistency constraints, introduces a strong inductive bias towards surface manifolds. This effectively regularizes optimization in sparse-view settings, suppressing floating artifacts and enhancing novel-view synthesis fidelity.

\subsubsection{Quantitative Evaluation of Reconstruction Process}
\label{subsubsec:quantitative_analysis_process}
Tables~\ref{tab:quality_eval_views} and~\ref{tab:efficiency_eval_views} provide a detailed quantitative breakdown of the reconstruction process for different methods across three distinct phases: initialization (2 views), exploration (10 views), and convergence (30 views). Note that NeRF-PRV~\cite{pan2024many} and MA-SCVP~\cite{pan2024integrating} are natively one-shot, auto-stop planners rather than the standard iterative NBV method. To align with our fixed-budget protocol, we evaluate these planners by sequentially executing their planned views up to the respective counts, carrying over final metrics if the planner terminates before reaching 30 views. Furthermore, we provide a comprehensive comparison between ObjSplat and these methods under their native auto-stop modes in Sec.~\ref{subsubsec:auto_stop_analysis}.

\textbf{Robust Initialization:} In the early stage ($N=2$), both our variants (NBV and NBP) significantly outperform baselines (FisherRF, GauSS-MI) in completeness (CR $\sim$40\% vs. $\sim$18\%) and geometric quality (F-Score $\sim$0.65 vs. $\sim$0.32). Similarly, NeRF-PRV exhibits lower novel-view synthesis quality, demonstrating limitations in extremely sparse views. The quantitative gap validates that the Gaussian surfel representation provides strong geometric priors, enabling robust reconstruction even with extremely sparse observations. In contrast, baselines struggle to infer correct geometry from limited views, resulting in higher depth errors (D-L1).

\textbf{The Efficiency-Completeness Trade-off:} At the exploration stage ($N=10$), we observe a distinct behavioral divergence between the greedy NBV and our NBP planner. Greedy NBV strategies generally exhibit strong novel-view synthesis capabilities (Test PSNR $>27$ dB) early on. Specifically, Ours-NBV variant achieves high completeness ($>90\%$) and rendering quality (>30dB) by aggressively transiting to distant, high-information viewpoints. This confirms that our geometry-aware uncertainty evaluation effectively identifies under-reconstructed regions. 
In contrast, the proposed NBP method adopts a more conservative yet efficient strategy, prioritizing path efficiency and local continuity. While achieving moderate coverage (74.41\%) at this stage, it does so with minimal movement (1.17 m). This indicates that the robot explores the local neighborhood thoroughly before transitioning to distant regions, avoiding the inefficient back-and-forth behavior typical of greedy planners.

\textbf{Global Optimality at Convergence:} Upon convergence ($N=30$), the advantages of the next-best-path planner become decisive. Both our variants achieve final coverage ($>91\%$), comparable to traditional point-cloud-based baselines, while significantly outperforming other GS-based baselines in completion ratio. Crucially, Ours-NBP achieves a final coverage and geometric accuracy comparable to the greedy NBV variant, but with a drastically reduced total path length (3.96 m vs. 18.02 m). This represents a 4$\times$ improvement in motion efficiency. Compared to the point-cloud-based PB-NBV~\cite{jia2025pb} (18.84 m), our method is nearly 4.7$\times$ more efficient, confirming that NBV strategies universally incur high motion overheads. Furthermore, while the appearance-side one-shot baseline, NeRF-PRV~\cite{pan2024many}, achieves competitive training fidelity, its novel-view and depth performance remain constrained by the slower convergence of NeRF-based representations in sparse-view settings. Additionally, its one-shot trajectory is determined by initial appearance priors, which may overlook fine-grained geometric details or self-occluded regions. Regarding reconstruction quality, Ours-NBP achieves the best or near-best performance across all metrics in the convergence stage. Notably, despite the conservative exploration strategy, NBP eventually surpasses greedy NBV in visual quality. This suggests that the smooth, continuous trajectory generated by NBP facilitates more consistent optimization of Gaussian surfels compared to the erratic viewpoints of greedy strategies, ultimately yielding a more coherent and photorealistic digital twin.

\begin{table*}[t!]
\centering
\caption{\textbf{Quantitative Evaluation of Reconstruction Quality.} We report visual quality (PSNR, SSIM, LPIPS) and geometric accuracy (Depth L1, Chamfer Distance, F-Score) on both training and novel test views.}
\label{tab:quality_eval_views}
\resizebox{1.0\textwidth}{!}{
\begin{tabular}{@{}l|c|
ccc ccc|  
ccc ccc|  
ccc ccc   
@{}}
\toprule

\textbf{Method} & \textbf{Split} & 
\multicolumn{6}{c|}{\textbf{Initialization (2 Views)}} &
\multicolumn{6}{c|}{\textbf{Exploration (10 Views)}} &
\multicolumn{6}{c}{\textbf{Convergence (30 Views)}} \\
\cmidrule(lr){3-8} \cmidrule(lr){9-14} \cmidrule(lr){15-20}

& &
\multicolumn{3}{c}{\textit{Visual Quality}} &
\multicolumn{3}{c|}{\textit{Geometry Quality}} &
\multicolumn{3}{c}{\textit{Visual Quality}} &
\multicolumn{3}{c|}{\textit{Geometry Quality}} &
\multicolumn{3}{c}{\textit{Visual Quality}} &
\multicolumn{3}{c}{\textit{Geometry Quality}} \\
\cmidrule(lr){3-8}
\cmidrule(lr){9-14}
\cmidrule(lr){15-20}

& &
PSNR$\uparrow$ & SSIM$\uparrow$ & LPIPS$\downarrow$ &
D-L1$\downarrow$ & CD$\downarrow$ & F-Score$\uparrow$ &
PSNR$\uparrow$ & SSIM$\uparrow$ & LPIPS$\downarrow$ &
D-L1$\downarrow$ & CD$\downarrow$ & F-Score$\uparrow$ &
PSNR$\uparrow$ & SSIM$\uparrow$ & LPIPS$\downarrow$ &
D-L1$\downarrow$ & CD$\downarrow$ & F-Score$\uparrow$
\\
\midrule

NeRF-PRV~\cite{pan2024many}
& \textit{Train}
& 42.53 & 0.991 & 0.015 & 95.784
& \multirow{2}{*}{-} & \multirow{2}{*}{-}
& 38.14 & 0.971 & 0.051 & 80.230
& \multirow{2}{*}{-} & \multirow{2}{*}{-}
& 34.39 & 0.947 & 0.095 & 51.294
& \multirow{2}{*}{-} & \multirow{2}{*}{-}
\\
& \textit{Test}
& 17.60 & 0.809 & 0.263 & 32.910 & &
& 20.24 & 0.852 & 0.209 & 47.195 & &
& 26.34 & 0.906 & 0.149 & 44.573 & &
\\ \midrule

FisherRF~\cite{jiang2024fisherrf}
& \textit{Train}
& 42.05 & 0.991 & 0.015 & 5.118
& \multirow{2}{*}{2.177} & \multirow{2}{*}{0.314}
& 35.73 & 0.965 & 0.067 & 1.939
& \multirow{2}{*}{2.196} & \multirow{2}{*}{0.531}
& 34.86 & 0.962 & 0.070 & 1.374
& \multirow{2}{*}{2.149} & \multirow{2}{*}{0.549}
\\
& \textit{Test}
& \textbf{25.43} & \textbf{0.921} & \underline{0.101} & 7.821 & &
& 28.02 & 0.941 & 0.086 & 3.255 & &
& 28.53 & 0.947 & 0.084 & 3.341 & &
\\ \midrule

GauSS-MI~\cite{xie2025gauss}
& \textit{Train}
& 42.64 & 0.991 & 0.014 & 3.842
& \multirow{2}{*}{2.169} & \multirow{2}{*}{0.322}
& 35.75 & 0.966 & 0.067 & 2.366
& \multirow{2}{*}{2.261} & \multirow{2}{*}{0.525}
& 35.08 & 0.964 & 0.069 & 1.514
& \multirow{2}{*}{2.191} & \multirow{2}{*}{0.590}
\\
& \textit{Test}
& \underline{24.13} & \underline{0.910} & 0.112 & 8.730 & &
& \underline{28.61} & \underline{0.943} & 0.083 & 3.181 & &
& 30.83 & 0.954 & 0.075 & 2.398 & &
\\ \midrule

Ours-NBV
& \textit{Train}
& \underline{43.59} & \underline{0.995} & \underline{0.009} & \textbf{0.741}
& \multirow{2}{*}{\textbf{1.660}} & \multirow{2}{*}{\underline{0.636}}
& \textbf{39.09} & \underline{0.989} & \underline{0.022} & \textbf{0.167}
& \multirow{2}{*}{\textbf{0.695}} & \multirow{2}{*}{\textbf{0.958}}
& \underline{36.66} & \underline{0.984} & \underline{0.028} & \textbf{0.153}
& \multirow{2}{*}{\textbf{0.609}} & \multirow{2}{*}{\textbf{0.973}}
\\
& \textit{Test}
& 22.39 & 0.909 & \textbf{0.095} & \textbf{2.856} & &
& \textbf{30.50} & \textbf{0.951} & \textbf{0.048} & \textbf{0.237} & &
& \underline{32.20} & \underline{0.965} & \underline{0.040} & \underline{0.191} & &
\\ \midrule

\textbf{Ours-NBP}
& \textit{Train}
& \textbf{43.65} & \textbf{0.996} & \textbf{0.006} & \underline{0.746}
& \multirow{2}{*}{\underline{1.835}} & \multirow{2}{*}{\textbf{0.653}}
& \underline{38.83} & \textbf{0.989} & \textbf{0.020} & \underline{0.212}
& \multirow{2}{*}{\underline{0.945}} & \multirow{2}{*}{\underline{0.875}}
& \textbf{36.76} & \textbf{0.985} & \textbf{0.028} & \underline{0.165}
& \multirow{2}{*}{\underline{0.611}} & \multirow{2}{*}{\underline{0.970}}
\\
& \textit{Test}
& 19.74 & 0.892 & 0.113 & \underline{4.919} & &
& 24.65 & 0.925 & \underline{0.080} & \underline{1.604} & &
& \textbf{32.35} & \textbf{0.966} & \textbf{0.039} & \textbf{0.189} & &
\\

\bottomrule
\end{tabular}}
\begin{tablenotes}
{\item[] \footnotesize
Comparison of visual (PSNR, SSIM, LPIPS) and geometric (Depth L1, Chamfer Distance, F-Score) quality across different view counts. Best values are shown in \textbf{bold}, and second-best values are \underline{underlined}.}
\end{tablenotes}
\vspace{-4mm}
\end{table*}

\begin{table*}[t!]
\centering
\caption{\textbf{Quantitative Analysis of Reconstruction Completeness and Efficiency.} We report reconstruction completion ratio (CR), completion (CE), and efficiency metrics (MC, online/offline time) at three distinct phases.}
\label{tab:efficiency_eval_views}
\resizebox{1.0\textwidth}{!}{
\begin{tabular}{@{}l|
ccccc| 
ccccc| 
ccccc  
@{}}
\toprule

\textbf{Method} &
\multicolumn{5}{c|}{\textbf{Initialization (2 Views)}} & \multicolumn{5}{c|}{\textbf{Exploration (10 Views)}} & \multicolumn{5}{c}{\textbf{Convergence (30 Views)}} \\
\cmidrule(lr){2-6} \cmidrule(lr){7-11} \cmidrule(lr){12-16}

& 
CR[\%]$\uparrow$ & CE[mm]$\downarrow$ &
MC[m]$\downarrow$ & Online[s]$\downarrow$ & Offline[s]$\downarrow$ &
CR[\%]$\uparrow$ & CE[mm]$\downarrow$ &
MC[m]$\downarrow$ & Online[s]$\downarrow$ & Offline[s]$\downarrow$ &
CR[\%]$\uparrow$ & CE[mm]$\downarrow$ &
MC[m]$\downarrow$ & Online[s]$\downarrow$ & Offline[s]$\downarrow$
\\
\midrule

SEE~\cite{border2024surface} & 57.04 & 8.88 & 0.53 & 20.03 & - & 72.47 & 4.88 & 1.65 & 172.91 & - & 89.14 & 1.71 & 4.32 & 420.80 & - \\
PB-NBV~\cite{jia2025pb}      & 57.88 & 9.88 & 0.37 & 18.66 & - & 87.27 & 2.63 & 5.65 & 82.22 & - & 90.69 & 2.04 & 18.84 & 223.44 & - \\
MA-SCVP~\cite{pan2024integrating} & 49.92 & 13.92 & 0.39 & 27.67 & - & 90.82 & 1.91 & 3.86 & 117.23 & - & 92.24 & 1.68 & 5.59 & 171.77 & - \\
\midrule

NeRF-PRV~\cite{pan2024many}
& - & - & 0.19 & \underline{14.66} & \textbf{33.37}
& - & - & \underline{1.87} & 81.28 & \textbf{36.56}
& - & - & \underline{5.78} & 207.08 & \textbf{38.55} \\

FisherRF~\cite{jiang2024fisherrf} 
& 19.45 & 32.17 & \underline{0.15} & 36.66 & 86.49 
& 63.98 & 6.22 & 3.49 & 246.95 & 97.54
& 69.72 & 5.05 & 11.74 & 732.10 & 101.07 \\

GauSS-MI~\cite{xie2025gauss} 
& 18.57 & 32.93 & \underline{0.15} & 21.07 & \underline{64.16}
& 66.53 & \underline{4.97} & 6.92 & 98.35 & \underline{86.27}
& 72.66 & 4.39 & 24.78 & 281.44 & \underline{91.51} \\

Ours-NBV
& \textbf{46.16} & \textbf{11.75} & 0.59 & 19.93 & 136.96 
& \textbf{90.04} & \textbf{2.47} & 6.42 & \underline{70.55} & 145.26 
& \textbf{91.89} & \textbf{2.04} & 18.02 & \underline{195.16} & 148.43 \\

\textbf{Ours-NBP} 
& \underline{41.35} & \underline{15.99} & \textbf{0.13} & \textbf{9.81} & 144.70 
& \underline{74.41} & 5.06 & \textbf{1.17} & \textbf{50.53} & 149.13 
& \underline{91.42} & \underline{2.11} & \textbf{3.96} & \textbf{121.96} & 146.21 \\

\bottomrule
\end{tabular}}
\begin{tablenotes}
{\item[] \footnotesize
Ours-NBP achieves comparable quality to the greedy NBV baseline but with significantly reduced motion and time costs. Best values are shown in \textbf{bold}, and second-best values are \underline{underlined}.
}
\end{tablenotes}
\vspace{-4mm}
\end{table*}

\textbf{System Efficiency:} Furthermore, our system demonstrates superior online efficiency. Benefiting from the localized updates of Gaussian surfels and the low movement cost inherent to progressive scanning, Ours-NBP requires only 121.96s for the entire online process—2.3$\times$ faster than GauSS-MI and 6$\times$ faster than FisherRF. Although some baselines may have lower offline refinement costs, our framework maintains a significant advantage in total mission time, capable of completing both online scanning and offline refinement within approximately 5 minutes. This confirms that ObjSplat is highly suitable for time-sensitive robotic applications. For a more detailed performance analysis, please refer to \cref{subsec:ablation}.

\subsubsection{Performance under Auto-Stop Mode}
\label{subsubsec:auto_stop_analysis}
While the fixed-budget analysis above illustrates the evolution of reconstruction, the system’s ability to autonomously determine completion is equally critical for practical deployment. Table \ref{tab:autostop_comparison} compares ObjSplat and the auto-stop planners~\cite{pan2024many, pan2024integrating} under their native termination criteria. Overall, Ours-NBP achieves a superior balance between reconstruction quality and efficiency. Specifically, Ours-NBP yields the highest PSNR (31.96 dB) while maintaining the lowest movement cost (3.55 m), outperforming NeRF-PRV in both visual fidelity and path length. The greedy NBV variant terminates earlier due to short-sighted evaluation, which triggers premature convergence and results in a slight drop in completeness. In contrast, Ours-NBP ensures more thorough surface coverage through the multi-step lookahead strategy, achieving results competitive with the point-cloud-based one-shot planner MA-SCVP. This confirms that our geometry-aware uncertainty provides a reliable termination signal, enabling high-fidelity reconstruction to be achieved autonomously with minimal motion costs.

\begin{table}[t!]
\centering
\caption{\textbf{Performance Comparison under Auto-Stop Mode.} We report the average number of views, reconstruction quality, and efficiency based on native termination criteria.}
\label{tab:autostop_comparison}
\resizebox{1.0\columnwidth}{!}{
\begin{tabular}{@{}l|c|ccccc@{}}
\toprule
\textbf{Method} & Views$\downarrow$ & PSNR[dB]$\uparrow$ & CD[mm]$\downarrow$ & CR[\%]$\uparrow$ & CE[mm]$\downarrow$ & MC[m]$\downarrow$ \\
\midrule
MA-SCVP~\cite{pan2024integrating} 
& \underline{15.62} & - & - & \textbf{92.24} & \textbf{1.68} & 5.59 \\ \midrule
NeRF-PRV~\cite{pan2024many} 
& 34.62 & 27.98 & - & - & - & 6.99 \\ 
Ours-NBV 
& \textbf{7.75} & \underline{29.73} & \underline{0.785} & 88.61 & 2.91 & \underline{5.09} \\
\textbf{Ours-NBP} 
& 25.44 & \textbf{31.96} & \textbf{0.644} & \underline{91.09} & \underline{2.19} & \textbf{3.55} \\
\bottomrule
\end{tabular}}
\vspace{-4mm}
\end{table}

\begin{figure*}[t]
    \centering
    \includegraphics[width=0.95\linewidth]{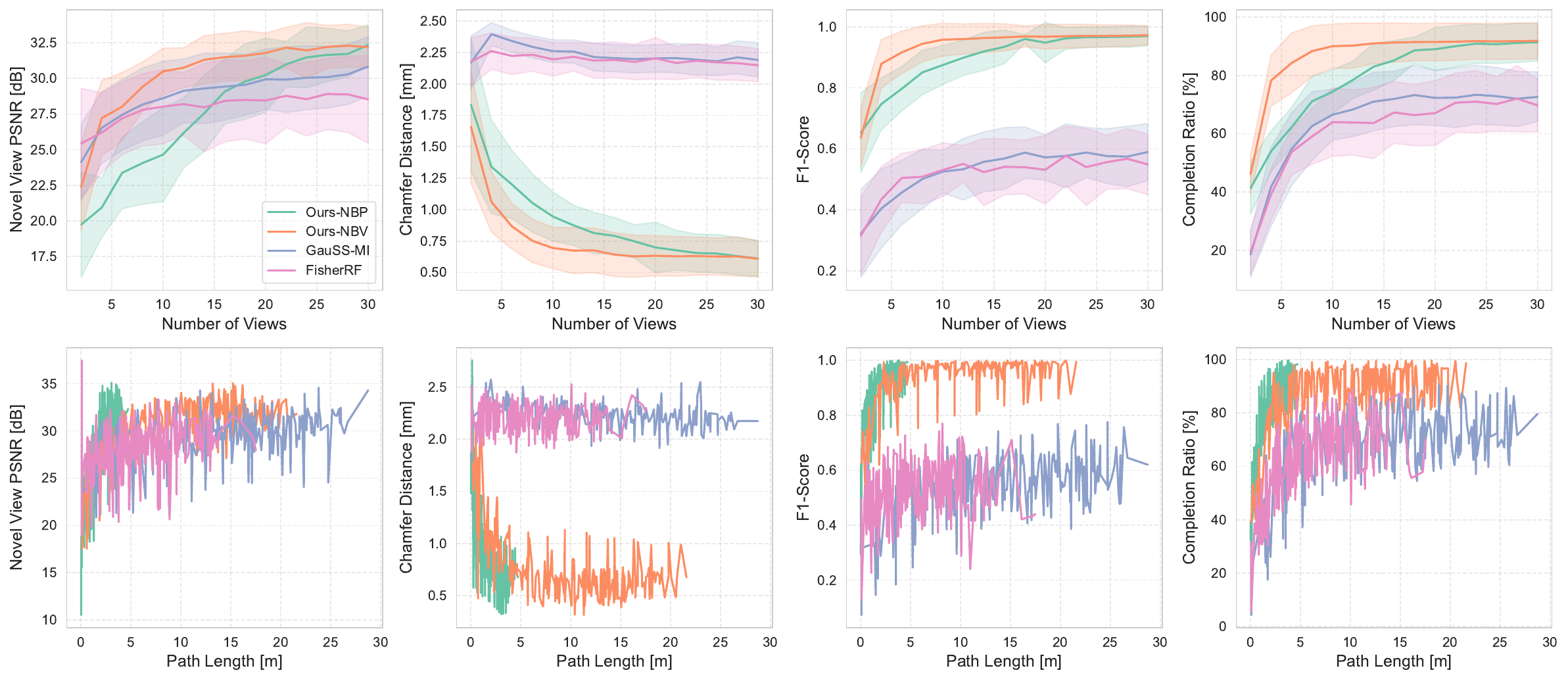}
    \caption{Quantitative comparison of reconstruction progress over the number of views (top row) and path length (bottom row). We report novel view PSNR, Chamfer Distance (CD), F1-Score, and Completion Ratio across 16 objects. Shaded areas indicate standard deviation.}
    \label{fig:qualitative_comparison_by_views_path}
\end{figure*}

\begin{figure*}[t]
    \centering
    \includegraphics[width=0.95\linewidth]{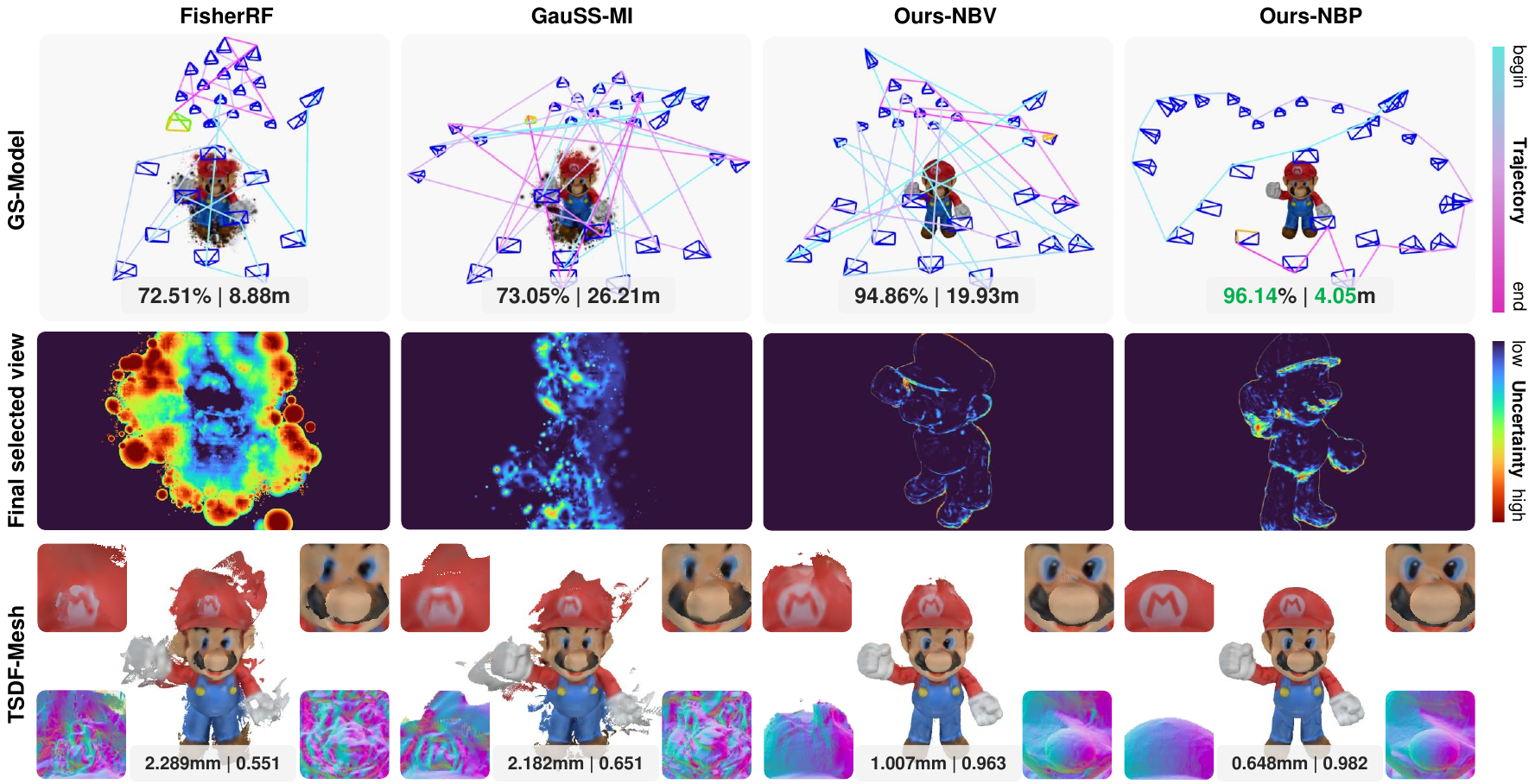}
    \caption{Visual comparison of reconstruction completeness and exploration efficiency on the \texttt{Mario} object. Top row: online Gaussian models with camera trajectories and frustums. Metrics denote Surface Coverage ($\uparrow$) $|$ Path Length ($\downarrow$). Middle row: the uncertainty map at the final selected viewpoint. Bottom row: final meshes extracted via TSDF fusion with zoomed-in details. Metrics denote Chamfer Distance ($\downarrow$) $|$ F-Score ($\uparrow$).}
    \label{fig:qualitative_comparison_traj}
    \vspace{-4mm}
\end{figure*}

\subsubsection{Reconstruction Process Analysis by Views vs. Path Length}
To further investigate the reconstruction dynamics, Fig.~\ref{fig:qualitative_comparison_by_views_path} plots the evolution of metrics against both the number of views and the accumulated movement cost. 
\textit{Convergence over Views (Top Row):}
In terms of view efficiency, Ours-NBV (orange) exhibits the fastest initial growth in PSNR and Completion Ratio, aligning with its greedy nature of prioritizing global information gain. However, Ours-NBP (green) demonstrates steady and robust convergence, catching up to NBV in all metrics by the end of the session (30 views). Furthermore, both variants demonstrate rapid geometric convergence: Chamfer Distance drops below 1.0 mm, and F1-Score exceeds 0.85 within the first 10 views. In contrast, baseline methods (GauSS-MI and FisherRF) plateau at a significantly higher error level ($>2.0$mm). This disparity underscores the geometric advantage of our Gaussian surfel representation in inferring accurate surface geometry from sparse observations.
\textit{Dominance in Path Efficiency (Bottom Row):}
The advantages of our NBP planner become irrefutable when metrics are plotted against movement cost. The Ours-NBP curves (green) exhibit a near-vertical ascent in the early phase, saturating at high quality levels (e.g., PSNR $>32$dB, Completion $>90\%$) within a minimal movement budget (about $5m$). In contrast, greedy Ours-NBV and baseline methods require trajectories extending beyond 15 meters to achieve comparable performance. This highlights that greedy strategies often incur high travel costs by jumping between distant informative viewpoints. Our NBP planner mitigates such redundant motions by thoroughly exploring local regions before transitioning, thereby maximizing information acquisition per unit of movement cost.
Furthermore, regarding geometric fidelity, both our variants maintain a significant lead over baselines throughout the entire process. Baseline curves remain relatively flat, indicating that simply adding views without geometric constraints fails to correct the underlying shape errors caused by overfitting. Additionally, our method exhibits smaller variance in novel view and reconstruction completeness across different objects, indicating superior adaptability and robustness to varying geometric and textural characteristics.

\begin{table}[t!]
\centering
\caption{\textbf{Ablation Study on Viewpoint Evaluation.} We evaluate the impact of each component on reconstruction quality and efficiency using the GSO dataset.}
\label{tab:ablation_view_eval}
\resizebox{0.95\columnwidth}{!}{
\begin{tabular}{@{}l|cccc@{}}
\toprule
\textbf{Method} & PSNR[dB]$\uparrow$ & CD[mm]$\downarrow$ & CR[\%]$\uparrow$ & MC[m]$\downarrow$ \\
\midrule
Opacity only & 28.75 & 0.701 & 86.02 & \textbf{3.84} \\
Confidence only & 29.99 & 0.693 & 86.95 & \underline{3.85} \\
\midrule
Ours w/o O.A. & \underline{32.22} & \underline{0.622} & \underline{91.07} & 4.09 \\
Ours w/o B.F. & 31.97 & 0.625 & 90.64 & 3.86 \\
Ours w/o Dyn. & 30.54 & 0.680 & 87.86 & 4.02 \\
\textbf{Ours} & \textbf{32.35} & \textbf{0.611} & \textbf{91.42} & 3.96 \\
\bottomrule
\end{tabular}}
\begin{tablenotes}
{\item[] \footnotesize
\textbf{O.A.}: Occlusion-Awareness; \textbf{B.F.}: Back-Face Observation; \textbf{Dyn.}: Dynamic Scheduling Strategy. Best results are \textbf{bold}, second best are \underline{underlined}.}
\end{tablenotes}
\vspace{-4mm}
\end{table}

\subsubsection{Qualitative Analysis of Reconstruction Process}
Fig.~\ref{fig:qualitative_comparison_traj} provides a qualitative breakdown of the reconstruction results on the \texttt{Mario} object, illustrating the trade-off between exploration efficiency and reconstruction quality. We visualize the online Gaussian models overlaid with camera trajectories (Row 1), uncertainty maps at the final selected viewpoint (Row 2), and the final extracted meshes (Row 3).

\textbf{Trajectory and Efficiency Analysis:}
As observed in the top row, baseline methods struggle to balance exploration with movement costs. GauSS-MI exhibits a highly chaotic trajectory (26.21m), oscillating between distant viewpoints due to the lack of global path constraints. FisherRF, while moving less, fails to explore the object comprehensively. Comparing our variants, Ours-NBV achieves high coverage but incurs a significant movement cost (19.93m), attributed to its greedy selection of spatially distant informative views. Conversely, Ours-NBP leverages the spatial topology graph to perform multi-step lookahead planning. This results in a smooth, enveloping trajectory that efficiently covers the object, achieving the highest coverage (96.14\%) with a drastically reduced path length (4.05m)—an improvement of nearly \textbf{5$\times$} in motion efficiency compared to the greedy NBV baseline. This reduction is driven by the occlusion and back-face awareness enabled by our surfel representation. Unlike prior approaches that rely on identifying unconverged areas to reduce uncertainty, our planner utilizes explicit normal information to infer unobserved rear surfaces, generating purposeful trajectories that distinguish necessary exploration from redundant movement.

\textbf{Reconstruction Quality and Uncertainty:}
Furthermore, baselines suffer from severe floating artifacts in the GS-Model and noise in the TSDF mesh, primarily caused by overfitting to sparse, ill-distributed views and a lack of object-centric foreground optimization. Their uncertainty maps (Row 2) often exhibit coarse-grained noise, failing to guide the robot effectively. 
In contrast, Ours-NBP produces a clean, water-tight mesh with sharp features (e.g., the hat emblem and face texture). The uncertainty visualization confirms that our geometry-aware metric correctly identifies fine-grained reconstruction quality and completeness cues—typically concentrated at object edges and occluded regions—thereby enabling targeted refinement. Consequently, Ours-NBP achieves the lowest Chamfer Distance (0.648mm) and the highest F-Score (0.982), demonstrating superior fidelity. Overall, our method generates a globally consistent, smooth trajectory, achieving the highest completeness and geometric accuracy with minimum movement cost.

\begin{table}[t!]
\centering
\caption{\textbf{Ablation Study on View Planning Strategies.} We compare different planning paradigms in terms of reconstruction quality and efficiency.}
\label{tab:ablation_view_plan}
\resizebox{0.95\columnwidth}{!}{
\begin{tabular}{@{}l|ccccc@{}}
\toprule
\textbf{Strategy} & PSNR $\uparrow$ & CD $\downarrow$ & CR $\uparrow$ & MC $\downarrow$ & Online $\downarrow$ \\
 & [dB] & [mm] & [\%] & [m] & [s] \\
\midrule
Random & 31.45 & 0.648 & 90.50 & 16.63 & 200.71 \\
Circle (Fixed) & 31.12 & 0.655 & 86.21 & \textbf{3.03} & \textbf{98.30} \\
\midrule
NBV (Greedy) & \underline{32.20} & \textbf{0.609} & \textbf{91.89} & 18.02 & 195.16 \\
NBV-1 + NBP & 31.62 & 0.652 & 89.12 & 4.51 & 130.18 \\
NBP-R (Step-wise) & 31.16 & 0.668 & 87.79 & 4.01 & 150.92 \\
\textbf{Ours (NBP-P)} & \textbf{32.35} & \underline{0.611} & \underline{91.42} & \underline{3.96} & \underline{121.96} \\
\bottomrule
\end{tabular}}
\begin{tablenotes}
{\item[] \footnotesize
\textbf{Circle}: efficient but blind; \textbf{NBV}: accurate but costly. Best results are \textbf{bold}, second best are \underline{underlined}.}
\end{tablenotes}
\vspace{-4mm}
\end{table}

\subsection{Ablation Study and System Performance}
\label{subsec:ablation}
To validate the design choices of ObjSplat, we conduct comprehensive ablation studies to assess the individual contributions of its core components.

\subsubsection{Ablation on Viewpoint Evaluation}
We first analyze the impact of different uncertainty quantification strategies on reconstruction results. We establish two baselines representing common heuristics: Opacity-only, which uses rendered opacity as a proxy for visibility, and Confidence-only, which employs raw surfel confidence without occlusion or back-face awareness. Subsequently,  we evaluated our method by selectively removing key components: Occlusion-Awareness (w/o O.A.), Back-Face check (w/o B.F.), and Dynamic Scheduling (w/o Dyn.). All variants employ the NBP planner to ensure a fair comparison. As shown in Table~\ref{tab:ablation_view_eval}, baseline heuristics (Opacity and Confidence) incur lower movement costs but suffer from significantly inferior completeness and quality. This indicates that in object-centric reconstruction, generic metrics are easily biased by the large proportion of empty background space, leading the system to prioritize trivial coverage over intricate surface details. As a result, these methods fail to detect critical geometric defects, such as self-occlusions or back-facing surfaces, resulting in premature termination. In contrast, our full approach incurs a slightly higher movement cost (3.96m). However, this additional motion is deliberate and goal-oriented, explicitly guided by our geometry-aware metric to refine under-reconstructed regions.

Further analysis reveals that removing any of the individual components causes varying degrees of degradation in reconstruction completeness and quality. Removing the occlusion-aware check causes erroneous co-visibility estimates due to self-occlusions, which results in overestimated confidence. Removing back-face observation hampers the identification of under-reconstructed surfaces in object-level tasks, where back-facing areas are non-closed surfaces. We note that the performance drops when removing the dynamic scheduling strategy, as this leads to degradation across both quality and completeness metrics. This demonstrates that, during the early stages of object reconstruction—when a large portion remains unconstructed—the system benefits from prioritizing exploration. As the model matures, the system shifts focus to refine finer under-reconstructed areas. By integrating these components, our method accurately identifies truly under-reconstructed regions, achieving an optimal balance between exploration and exploitation.

\begin{table}[t!]
\centering
\caption{\textbf{Ablation on Unified Reconstruction Backends.} Evaluating identical 30-view trajectories under different mappers (3DGS vs. GSurfels) to isolate mapper and planner contributions.}
\label{tab:unified_backend_comparison}
\resizebox{0.95\columnwidth}{!}{
\begin{tabular}{@{}l|c|ccccc@{}}
\toprule
\textbf{Method} & \textbf{Mapper} & PSNR[dB]$\uparrow$ & LPIPS$\downarrow$ & CD[mm]$\downarrow$ & CR[\%]$\uparrow$ & MC[m]$\downarrow$ \\
\midrule
FisherRF~\cite{jiang2024fisherrf}  & MonoGS & 29.01 & 0.082 & 2.166 & 72.96 & \underline{12.51} \\
GauSS-MI~\cite{xie2025gauss}  & (3DGS) & 30.16 & 0.078 & 2.153 & 73.59 & 24.88 \\
\midrule
FisherRF~\cite{jiang2024fisherrf}  & \multirow{3}{*}{ObjSplat} & 30.23 & 0.049 & 0.758 & 89.79 & \underline{12.51} \\
GauSS-MI~\cite{xie2025gauss}  & \multirow{3}{*}{(GSurfels)} & 31.54 & 0.041 & 0.685 & 90.49 & 24.88 \\
Ours-NBV  & & \underline{32.20} & \underline{0.040} & \textbf{0.609} & \textbf{91.89} & 18.02 \\
\textbf{Ours-NBP} & & \textbf{32.35} & \textbf{0.039} & \underline{0.611} & \underline{91.42} & \textbf{3.96} \\
\bottomrule
\end{tabular}}
\end{table}

\begin{table}[t!]
\centering
\caption{\textbf{Parameter Sensitivity Analysis.} We evaluate the impact of key hyperparameters on the reconstruction result.}
\label{tab:ablation_parameters}
\resizebox{0.95\columnwidth}{!}{
\begin{tabular}{@{}l|c|c|cccc@{}}
    \toprule
    \textbf{Parameter} & \textbf{Value} & Views & PSNR[dB]$\uparrow$ & CD[mm]$\downarrow$ & CR[\%]$\uparrow$ & MC[m]$\downarrow$ \\
    \midrule
    \textbf{Default Setup} & -- & 30 & \underline{32.35} & 0.611 & 91.42 & 3.96 \\
    \midrule
    \multirow{2}{*}{\begin{tabular}[l]{@{}l@{}}
    Texture Error \\
    Thresh. ($\tau_C$)
    \end{tabular}}
    & 0.15 & 30 & 32.27 & 0.615 & 91.42 & 3.98 \\
    & 0.35 & 30 & 32.29 & 0.617 & 91.41 & 4.01 \\
    \midrule
    \multirow{2}{*}{\begin{tabular}[l]{@{}l@{}}
    Depth Tolerance \\
    Thresh. ($\tau_d$)
    \end{tabular}}
    & 0.003 & 30 & \textbf{32.46} & 0.616 & 91.29 & 4.08 \\
    & 0.007 & 30 & 32.27 & 0.627 & 91.37 & 4.00 \\
    \midrule
    \multirow{2}{*}{\begin{tabular}[l]{@{}l@{}}
    Local Window \\
    Size ($k+1$)
    \end{tabular}}
    & 5  & 30 & 32.11 & 0.622 & \textbf{91.59} & 4.04 \\
    & 15 & 30 & 32.05 & 0.614 & 91.21 & 4.12 \\
    \midrule
    \multirow{2}{*}{\begin{tabular}[l]{@{}l@{}}
    NBP Weight \\
    ($\lambda$ in Eq.~\ref{eq:path_score})
    \end{tabular}}
    & 0.2 & 30 & 32.20 & \underline{0.610} & 91.24 & 3.90 \\
    & 0.8 & 30 & 32.27 & \textbf{0.605} & \underline{91.50} & 4.01 \\
    \midrule
    \multirow{2}{*}{\begin{tabular}[l]{@{}l@{}}
    Auto Stop \\
    Thresh. ($\tau_{\text{stop}}$)
    \end{tabular}}
    & 0.02 & 27.31 & 31.75 & 0.621 & 91.42 & \underline{3.82} \\
    & 0.1 & 21.94 & 30.78 & 0.685 & 89.31 & \textbf{3.08} \\
\bottomrule
\end{tabular}}
\vspace{-4mm}
\end{table}

\subsubsection{Ablation on View Planning Strategy}
Table~\ref{tab:ablation_view_plan} evaluates distinct planning paradigms. The Random strategy, which randomly samples viewpoints distributed uniformly across the hemisphere, is effective for covering most of the object’s surface. This resilience is attributed to our Gaussian surfel representation and local visibility-based optimization in handling non-sequential data. However, the fixed-trajectory Circle baseline, while minimizing movement and time, yields poor reconstruction fidelity. This confirms that dynamic, scene-responsive active perception is essential for capturing complex geometric details missed by fixed paths. Conversely, the greedy NBV strategy achieves high completeness and geometric accuracy but incurs excessive movement costs, rendering it impractical for time-sensitive autonomous tasks.

A key finding is the superiority of executing the entire planned path (Ours / NBP-P) over the receding horizon approach (NBP-R), where only the first step is executed before replanning. NBP-R performs significantly worse, likely due to its overly conservative nature, which hinders extensive exploration within a limited view budget. Furthermore, frequent replanning increases computational overhead and disrupts global scanning objectives. We also evaluated a hybrid strategy (NBV-1 + NBP) that initializes with a single greedy view. However, the initial large displacement disrupts trajectory smoothness, leading to suboptimal results.

Ultimately, Ours (NBP-P) achieves the best trade-off: it matches the high reconstruction fidelity of the computationally expensive NBV strategy while maintaining a low movement cost comparable to fixed trajectories. Moreover, it achieves the lowest online processing time among active methods, as the reduced planning frequency allows more computational resources to be allocated to real-time mapping and optimization.

\begin{table}[t!]
\centering
\caption{\textbf{Ablation Study on Geometry-Texture Optimization.} We evaluate the contribution of each component by removing individual loss terms from the optimization process.}
\label{tab:ablation_gs_optim}
\resizebox{0.95\columnwidth}{!}{
\begin{tabular}{@{}l|ccccc@{}}
\toprule
\textbf{Method} & PSNR[dB]$\uparrow$ & SSIM$\uparrow$ & LPIPS$\downarrow$ & CD[mm]$\downarrow$ & CR[\%]$\uparrow$ \\
\midrule
Ours w/o $\mathcal{L}_n$ & 32.01 & 0.961 & 0.046 & 0.750 & 88.23 \\
Ours w/o $\mathcal{L}_c$ & 32.31 & 0.963 & \underline{0.040} & 0.613 & \textbf{91.52} \\
Ours w/o $\mathcal{L}_m$ & \underline{32.33} & 0.963 & 0.041 & 0.612 & 90.64 \\
Ours w/o $\mathcal{L}_o$ & 32.25 & \underline{0.965} & 0.042 & \textbf{0.608} & 90.98\\
\textbf{Ours} & \textbf{32.35} & \textbf{0.966} & \textbf{0.039} & \underline{0.611} & \underline{91.42} \\
\bottomrule
\end{tabular}}
\end{table}

\begin{figure}[t!]
    \centering
    \includegraphics[width=0.85\linewidth]{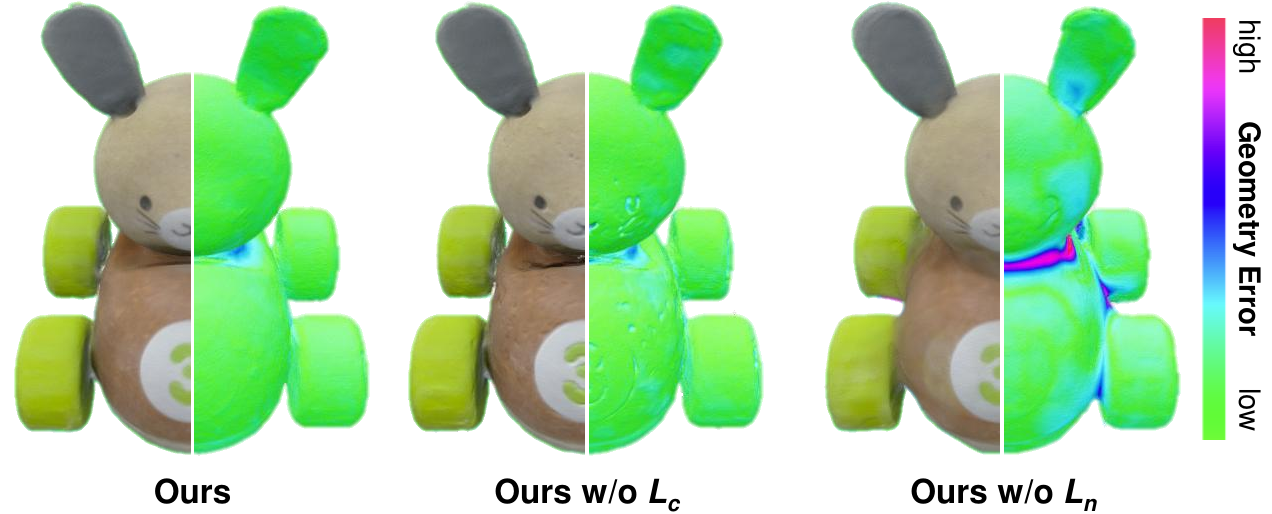}
    \caption{Reconstruction quality comparison on the \texttt{Bunny Racer} object.}
    \label{fig:ablation_opt}
    \vspace{-4mm}
\end{figure}

\subsubsection{Ablation on Unified Reconstruction Backend}
To isolate the contribution of our planning strategy from the underlying object representation, we conduct an additional controlled experiment to evaluate 3DGS-based planners under a unified Gaussian Surfel backend (see Table~\ref{tab:unified_backend_comparison}). While switching to the GSurfels mapper consistently improves the baseline performance (e.g., GauSS-MI’s PSNR increases by 1.38 dB), our methods still achieve higher reconstruction fidelity and geometric accuracy. Specifically, Ours-NBP outperforms GauSS-MI by 0.81 dB in PSNR and 0.074 mm in CD, demonstrating that our geometry-aware evaluation identifies more informative viewpoints that are essential for resolving complex geometric ambiguities. Notably, Ours-NBP achieves superior quality while reducing movement cost by over 84\% compared to GauSS-MI (3.96m vs. 24.88m). This confirms that while the GSurfel representation raises the performance ceiling, the core gains in exploration efficiency and final completeness are primarily driven by our proposed geometry-aware view evaluation and NBP planning strategy.

\begin{figure*}[t]
    \centering
    \includegraphics[width=0.95\linewidth]{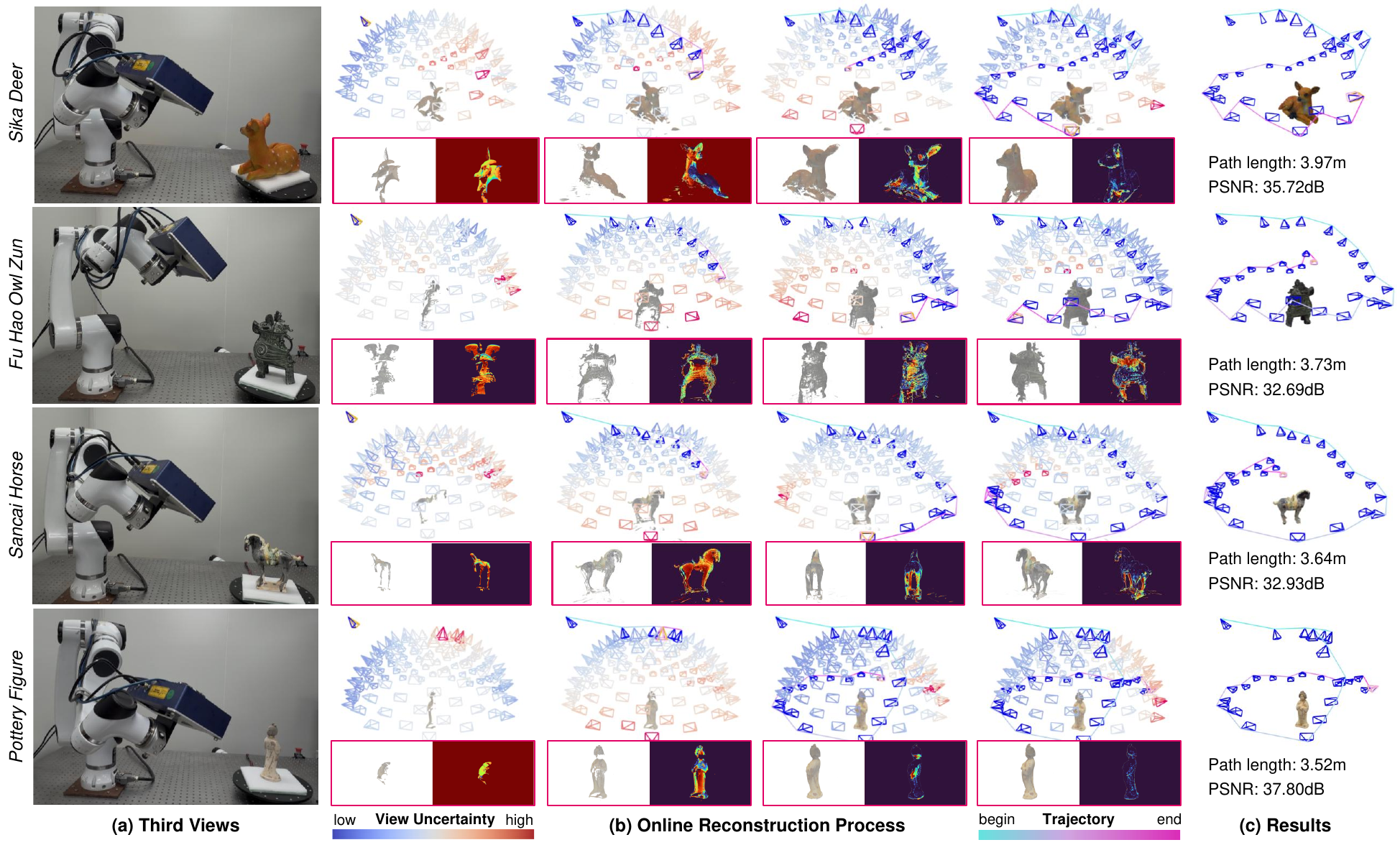}
    \caption{\textbf{Real-World Progressive Reconstruction Process Results.} We show four snapshots of the NBP planning execution. In each step, the robot executes a path to a sub-goal with high estimated uncertainty (visualized in the inset). The final column shows the complete trajectory, total path length, and the final PSNR (dB) of the training views.}
    \label{fig:realworld_recon_process}
    \vspace{-4mm}
\end{figure*}

\begin{figure*}[t]
    \centering
    \includegraphics[width=0.95\linewidth]{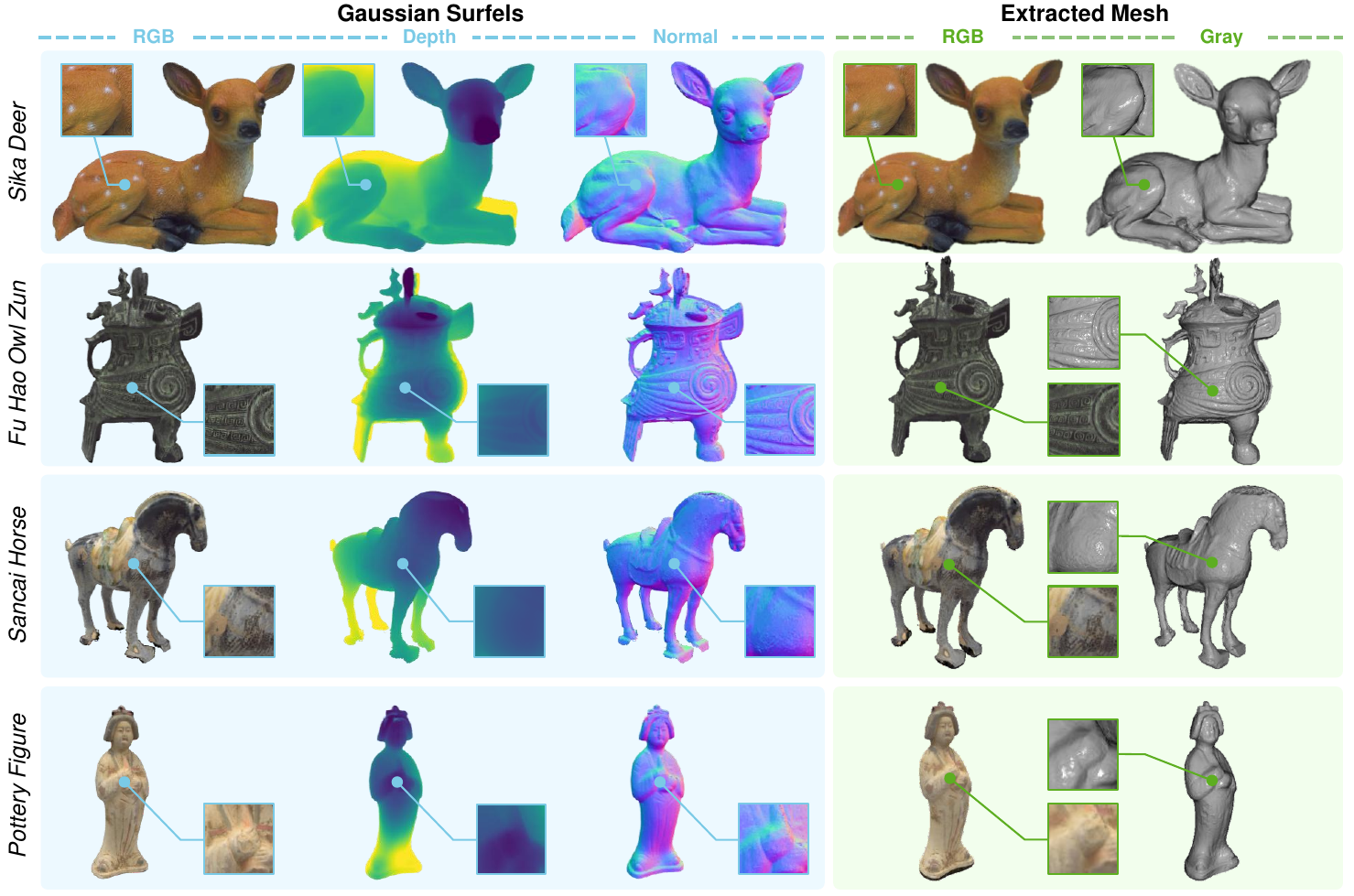}
    \caption{\textbf{Final reconstruction results of real-world artifacts.} We display the rendered images from the final Gaussian surfels (left) and the corresponding extracted surface meshes (right). Our method recovers high-fidelity texture and geometry even for objects with complex topology and surface patterns.}
    \label{fig:realworld_recon_results}
\end{figure*}

\begin{table}[t!]
\centering
\caption{Impact of the object's initial pose. (30 views and 5 trials)}
\label{tab:ablation_initial_pose}
\resizebox{1.0\columnwidth}{!}{
\begin{tabular}{@{}l|cccc@{}}
\toprule
\textbf{Object} & PSNR[dB]$\uparrow$ & CD[$mm$]$\downarrow$ & CR[\%]$\uparrow$ & MC[$m$]$\downarrow$ \\
\midrule
\texttt{Bunny} 
    & 32.68 $\pm$ 1.02
    & 0.718 $\pm$ 0.016
    & 87.16 $\pm$ 0.29
    & 3.89 $\pm$ 0.13 \\

\texttt{Chicken} 
    & 32.14 $\pm$ 0.61
    & 0.598 $\pm$ 0.010
    & 86.85 $\pm$ 0.46
    & 4.06 $\pm$ 0.44 \\

\texttt{Turtle} 
    & 31.26 $\pm$ 1.74
    & 0.747 $\pm$ \underline{0.006}
    & 84.28 $\pm$ 0.39
    & 3.94 $\pm$ 0.16 \\

\texttt{Elephant} 
    & 31.98 $\pm$ 0.87
    & 0.730 $\pm$ 0.019
    & 92.30 $\pm$ 0.37
    & 4.24 $\pm$ 0.19 \\

\texttt{Ortho} 
    & 33.48 $\pm$ \textbf{0.37}
    & 0.642 $\pm$ 0.019
    & 80.83 $\pm$ 1.04
    & 4.08 $\pm$ \underline{0.10} \\

\texttt{Horse} 
    & 33.92 $\pm$ 0.74
    & 0.477 $\pm$ 0.025
    & \textbf{99.54} $\pm$ 0.28
    & 4.05 $\pm$ 0.27 \\

\texttt{Eagle} 
    & 29.76 $\pm$ 1.32
    & 0.610 $\pm$ 0.010
    & 96.51 $\pm$ 1.19
    & 4.20 $\pm$ 0.24 \\

\texttt{Dragon} 
    & 31.02 $\pm$ 0.52
    & 0.746 $\pm$ 0.033
    & 95.77 $\pm$ 0.53
    & 4.13 $\pm$ 0.19 \\

\texttt{Mario} 
    & 32.68 $\pm$ 0.82
    & 0.650 $\pm$ 0.041
    & 96.19 $\pm$ 0.25
    & 4.25 $\pm$ 0.30 \\

\texttt{Yoshi} 
    & 32.48 $\pm$ \underline{0.48}
    & 0.503 $\pm$ 0.012
    & 96.40 $\pm$ 0.34
    & 3.88 $\pm$ 0.13 \\

\texttt{Dino} 
    & 32.25 $\pm$ 0.60
    & 0.683 $\pm$ 0.029
    & \underline{96.94} $\pm$ 0.82
    & 4.26 $\pm$ 0.21 \\

\texttt{Backpack} 
    & 29.49 $\pm$ 0.85
    & 0.966 $\pm$ 0.037
    & 80.83 $\pm$ 0.62
    & 4.31 $\pm$ 0.29 \\

\texttt{Shoe} 
    & 31.51 $\pm$ 1.21
    & 0.555 $\pm$ 0.010
    & 82.77 $\pm$ \underline{0.10}
    & \textbf{3.66} $\pm$ 0.26 \\

\texttt{Candy-box} 
    & \underline{34.12} $\pm$ 1.07
    & \textbf{0.378} $\pm$ 0.079
    & 94.80 $\pm$ \textbf{0.05}
    & \underline{3.75} $\pm$ 0.27 \\

\texttt{Cup} 
    & 32.84 $\pm$ 1.16
    & 0.485 $\pm$ 0.018
    & 96.51 $\pm$ 0.16
    & 4.01 $\pm$ 0.16 \\

\texttt{Stacking} 
    & \textbf{34.74} $\pm$ 1.10
    & \underline{0.449} $\pm$ \textbf{0.005}
    & 87.15 $\pm$ 0.15
    & 4.01 $\pm$ \textbf{0.08} \\

\midrule
\textbf{Avg.} & 32.27 $\pm$ 1.45 & 0.621 $\pm$ 0.149 & 90.93 $\pm$ 6.45 & 4.05 $\pm$ 0.18 \\
\bottomrule
\end{tabular}}
\vspace{-4mm}
\end{table}

\subsubsection{Parameter Sensitivity Analysis}
\label{subsubsec:param_analysis}
To evaluate system reliability, we analyze its sensitivity to five key hyperparameters (see Table~\ref{tab:ablation_parameters}). Overall, varying the threshold parameters leads to only minor performance changes, demonstrating our system's inherent robustness without careful parameter tuning. Decreasing $k$ slightly weakens multi-view constraints, while a larger $k$ may introduce views from overly distant or oblique views, leading to degraded appearance quality. We empirically find that $k\!=\!9$ strikes a good balance between enforcing multi-view consistency and avoiding noisy or less informative observations. The parameter $\lambda$ controls the trade-off between maximizing information gain and minimizing movement cost in the NBP planner. Assigning a higher weight to information gain generally improves reconstruction quality, at the cost of longer paths, reflecting the exploration--efficiency trade-off in active perception. Furthermore, the auto-stop threshold ($\tau_{\text{stop}}$) explicitly controls the adaptive termination point. A stricter threshold requires more views to achieve higher completeness, while a looser one enables earlier termination (averaging $\sim$22 views) with minor quality degradation, confirming our uncertainty metric is a reliable indicator for autonomous stopping.

\subsubsection{Ablation on Geometry-Texture Joint Optimization}
We selectively remove different loss terms during the optimization process to perform a quantitative analysis (see Table~\ref{tab:ablation_gs_optim}). The normal consistency loss helps reduce geometric fluctuations caused by overfitting to texture edges (see Fig.~\ref{fig:ablation_opt}), which even leads to inflated completeness. Interestingly, removing opacity regularization leads to an improvement in CD. We hypothesize that the strict binary opacity constraint may hinder the accurate modeling of complex self-occluded regions, whereas relaxing this constraint allows for better geometric fitting at the cost of visual quality. By balancing trade-offs, we achieve a favorable compromise between geometric accuracy and visual fidelity.

\begin{table}[!t]
    \caption{Average Processing time per step}
    \centering
    \resizebox{\linewidth}{!}{
    \begin{tabular}{ccccc}
        \toprule
        \textbf{Mapper} & \textbf{Candidate views} & \textbf{Get uncertainty} & \textbf{Path planning} & \textbf{Visualizer (optional)} \\
        \hline
        326.40ms & 2.71ms & 164.78ms & 21.91ms & 94.28ms \\ 
        \midrule
        \textbf{Segmentation}  & \textbf{Extract Obj-PCD} & \textbf{Move turntable} & \textbf{Move arm} & \textbf{} \\
        \hline
        361.68ms & 51.25ms & 1399.40ms & 1461.62ms & \\
        \bottomrule 
    \end{tabular}
    }
    \label{tab:system_performance}
\vspace{-4mm}
\end{table}

\subsubsection{Robustness to Initial Pose}
To evaluate robustness to the initial object orientation, we randomly perturb the object pose around the z-axis. For each of the 16 GSO objects, we sample 5 random initial yaw angles and evaluate the final reconstruction after 30 views. As shown in Table~\ref{tab:ablation_initial_pose}, the system demonstrates consistent robustness, with low standard deviations across metrics. This confirms that our NBP planner and geometry-aware uncertainty effectively guide the robot to explore critical regions regardless of the starting configuration, ensuring reliable high-fidelity reconstruction.

\subsubsection{Runtime Analysis}
Table~\ref{tab:system_performance} reports the average processing time per step for different modules. The system is capable of completing a full perception-planning-action loop in approximately 4 seconds. Benefiting from the sparse decision-making within the spatial topology and efficient rendering-based view evaluation enabled by the unified Gaussian surfel representation, the computational overhead for view evaluation and path planning remains minimal, constituting only a small fraction of the total processing load.

\subsection{Real-world Experiments}
\label{subsec:real_world_exp}
To validate the effectiveness of ObjSplat in real-world scenarios, we deploy the system on the robotic platform depicted in Fig.~\ref{fig:experimental_platforms}. To ensure coherent model fusion amidst mechanical vibration and calibration noise, we activated the object-centric pose tracking pipeline (described in Sec.~\ref{subsubsec:object_centric_tracking}) throughout all experiments, maintaining global coordinate consistency. Fig.~\ref{fig:realworld_recon_process} visualizes the autonomous reconstruction process for four physical objects, displaying four snapshots of the NBP planning process. In each slice, the uncertainty map rendered from a selected sub-goal viewpoint is shown alongside the executed trajectory. The system actively identifies high-uncertainty regions (red/bright areas) and generates collision-free paths for exploration. As reconstruction progresses, the global surface uncertainty diminishes, eventually confining residual uncertainty to deeply occluded regions or complex undersides that are difficult to observe. Quantitatively, our system completed the reconstruction of these complex objects with an average trajectory length of less than 4.0 m, confirming that the NBP planner effectively translates its "lookahead" capability into highly efficient, non-redundant scanning trajectories.

Fig.~\ref{fig:realworld_recon_results} presents the final reconstruction results, displaying both optimized Gaussian surfel models and extracted meshes. The results demonstrate that ObjSplat achieves accurate geometry and photorealistic texture fidelity across objects with diverse characteristics. Specifically, it faithfully recovers fine-grained details, such as the skin texture of the \texttt{Sika Deer}, intricate geometric carvings on the \texttt{Fu Hao Owl Zun}, and rich patterns on the \texttt{Pottery Figure}. The explicit mesh extraction further validates the geometric precision, exhibiting smooth surfaces with sharp features. Crucially, these high-fidelity digital assets are directly applicable to downstream physical simulation tasks (see Fig.~\ref{fig: teaser}). For a complete visualization of the scanning process and $360^\circ$ renderings, please refer to the supplementary video.

%% file: src/5_conclusion.tex
\section{Conclusion and Future Work}
\label{sec:conclusion_future_work}

In this paper, we presented \ours, a unified active object reconstruction framework that bridges the gap between high-fidelity digitization and efficient robotic exploration. By adopting 2D Gaussian surfels as a shared representation, our system tightly couples incremental model updates with geometry-aware view planning. We introduce a geometry-aware view evaluation pipeline that explicitly accounts for back-face visibility and multi-view covisibility, providing reliable guidance for targeted refinement beyond simple opacity cues. To overcome the short-sightedness of conventional greedy strategies, our next-best-path planner performs multi-step lookahead on a dynamic spatial topology, effectively optimizing the trade-off between information gain and movement cost. Extensive evaluations in both simulation and real-world setups demonstrate that \ours consistently outperforms state-of-the-art methods, producing photorealistic, watertight digital assets with significantly reduced autonomous scanning time.

Despite these advancements, our system has limitations that open avenues for future research. First, the current framework is tailored for the reconstruction of single, static, rigid objects. Handling complex optical properties, such as extreme lighting variations, transparency, or high specularity, remains challenging for the current rendering model. Future work will focus on integrating robust material estimation techniques and exploring dynamic object representations, broadening the scope of autonomous digitization to interactive and uncontrolled real-world environments.